%% file: neurips_2025.tex
\documentclass{article}

\usepackage[preprint]{neurips_2025}

\usepackage[utf8]{inputenc} 
\usepackage[T1]{fontenc}    
\usepackage{hyperref}       
\usepackage{url}            
\usepackage{booktabs}       
\usepackage{amsfonts}       
\usepackage{nicefrac}       
\usepackage{microtype}      
\usepackage{xcolor}         
\usepackage{enumitem} 
\usepackage{multirow}
\usepackage{makecell}
\usepackage{graphicx}
\usepackage{caption}
\usepackage{pifont}
\usepackage{subfigure}
\usepackage{amsthm}
\usepackage{wrapfig}
\usepackage{amsmath}
\usepackage{algorithm}
\usepackage{algorithmic}
\usepackage{hyperref}
\usepackage[textsize=tiny]{todonotes}
\usepackage{adjustbox}

\usepackage{comment}

\theoremstyle{plain}
\newtheorem{theorem}{Theorem}[section]

\theoremstyle{definition}

\newtheorem{assumption}{Assumption}[section]
\theoremstyle{remark}

\title{Decouple and Orthogonalize: \\ A Data-Free Framework for LoRA Merging}

\author{%
  Shenghe Zheng$^{1,2}$, Hongzhi Wang$^{1}$, Chenyu Huang$^{3}$, Xiaohui Wang$^{3}$, Tao Chen$^{3}$, \\ \textbf{Jiayuan Fan}$^{3}$, \textbf{Shuyue Hu}$^{2}$, \textbf{Peng Ye}$^{2,4}$ \\
  $^1$ Harbin Institute of Technology \quad $^2$ Shanghai AI Laboratory $^3$ Fudan University \quad  \\ $^4$ The Chinese University of Hong Kong \\
  \texttt{shenghez.zheng@gmail.com} \\
}

\begin{document}

\maketitle

\input{Sections/Abstract}
\input{Sections/Introduction}
\input{Sections/Related_Work}
\input{Sections/Method}

\input{Sections/Experiments}
\input{Sections/Conclusion}

\bibliographystyle{plain}
\bibliography{neurips_2025}

\newpage
\appendix
\input{Sections/Appendix}



\end{document}

%% file: Sections/Abstract.tex
\begin{abstract}
With more open-source models available for diverse tasks, model merging has gained attention by combining models into one, reducing training, storage, and inference costs.  Current research mainly focuses on model merging for full fine-tuning, overlooking the popular LoRA. However, our empirical analysis reveals that:
a) existing merging methods designed for full fine-tuning perform poorly on LoRA; b) LoRA modules show much larger parameter magnitude variance than full fine-tuned weights; c) greater parameter magnitude variance correlates with worse merging performance. 
Considering that large magnitude variances cause deviations in the distribution of the merged parameters, resulting in information loss and performance degradation,
we propose a \textbf{D}ecoupled and \textbf{O}rthogonal merging approach~(\textbf{DO-Merging}). 
By separating parameters into magnitude and direction components and merging them independently, we reduce the impact of magnitude differences on the directional alignment of the merged models, thereby preserving task information. Furthermore, we introduce a data-free, layer-wise gradient descent method with orthogonal constraints to mitigate interference during the merging of direction components. We provide theoretical guarantees for both the decoupling and orthogonal components. And we validate through extensive experiments across vision, language, and multi-modal domains that our proposed DO-Merging can achieve significantly higher performance than existing merging methods at a minimal cost. Notably, each component can be flexibly integrated with existing methods, offering near free-lunch improvements across tasks.
\end{abstract}

%% file: Sections/Introduction.tex
\section{Introduction}

Deep learning is widely used in many applications~\cite{devlin-etal-2019-bert,touvron2023llama}. However, edge-side users often lack strong resources or large datasets, and thus prefer ready-to-use models tailored to their specific tasks. The rapid growth of open-source platforms like HuggingFace~\cite{wolf2019huggingface} has made this goal increasingly achievable. In practice, real-world tasks often involve multiple subtasks~\cite{yangadamerging,jin2022dataless}. Handling each with a separate model increases cost and deployment complexity. Model merging addresses this by combining existing models into a single model capable of handling all target tasks, reducing both retraining and deployment costs~\cite{ilharco2022editing,matena2022merging,yadav2024ties}. This approach has recently gained great attention.

Current research on model merging mainly targets task interference. Methods are categorized by their target models: full fine-tuning or PEFT techniques like LoRA~\cite{hu2022lora}. For full fine-tuning, merging approaches fall into three types: automatic coefficient computation~\cite{yangadamerging,jin2022dataless,matena2022merging}, optimization-based conflict reduction using task vectors~\cite{yadav2024ties,davari2023model,du2024parameter,free}, and dedicated modules for task-specific knowledge~\cite{huang2024emr,lu2024twin}. However, these methods 
may not be suitable for LoRA,
or require architectural changes, limiting usability. Thus, specialized LoRA merging techniques are needed. 
Existing LoRA-specific methods mainly include subspace projection merging~\cite{knots,lego} and parameter-driven coefficient computation~\cite{iteris,klora,lorasoups,lorahub}. Yet, they often result in limited gains or lack generalization across tasks. Therefore, new and more effective merging strategies are necessary.

\begin{figure*}[t]
	\centering
	\subfigure[]{
		\begin{minipage}[t]{0.32\linewidth}
			\centering
	\includegraphics[width=1\textwidth]{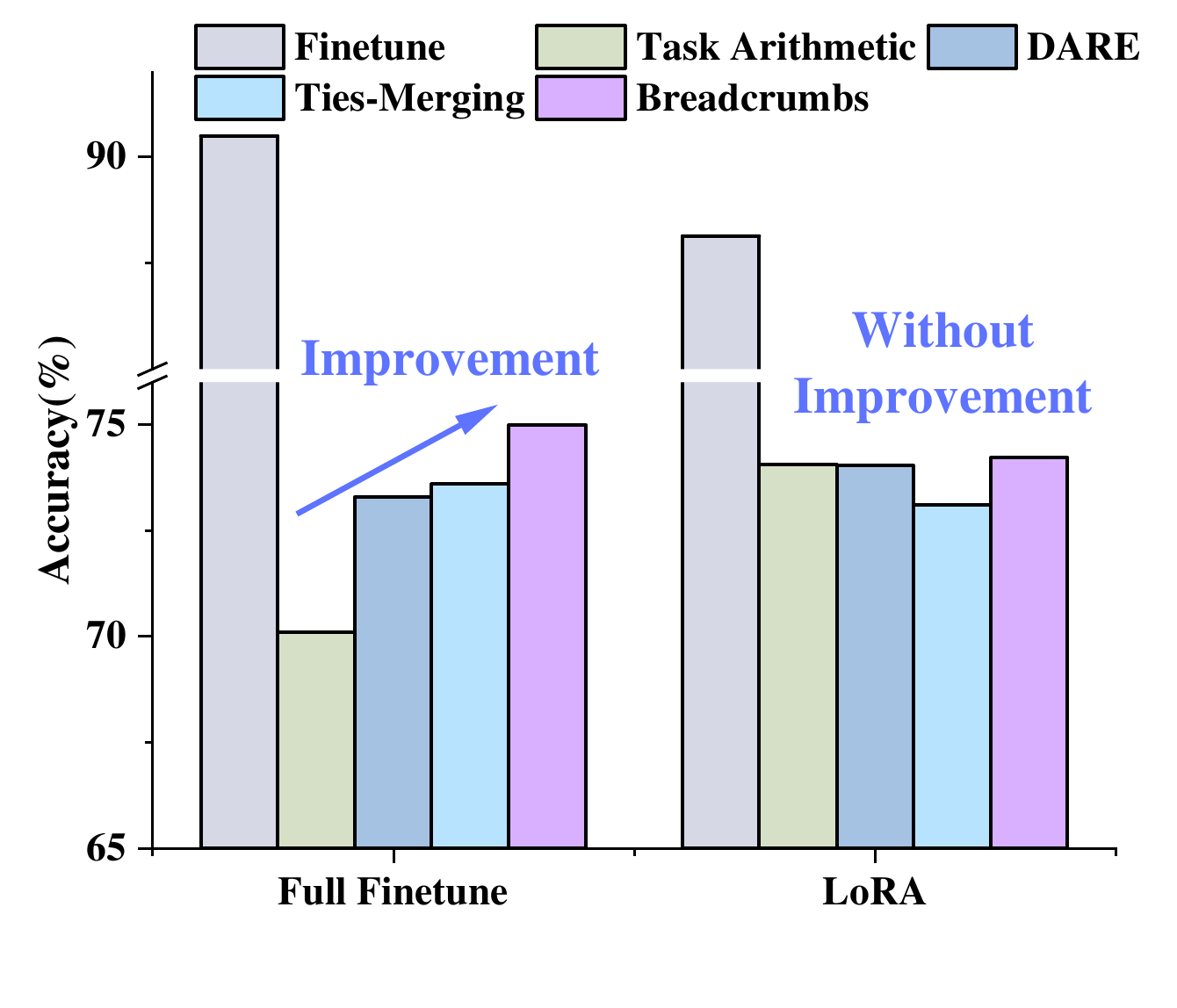}
		\end{minipage}
        \vspace{-0.3cm}
	}%
	\subfigure[]{
		\begin{minipage}[t]{0.32\linewidth}
			\centering
	\includegraphics[width=1\textwidth]{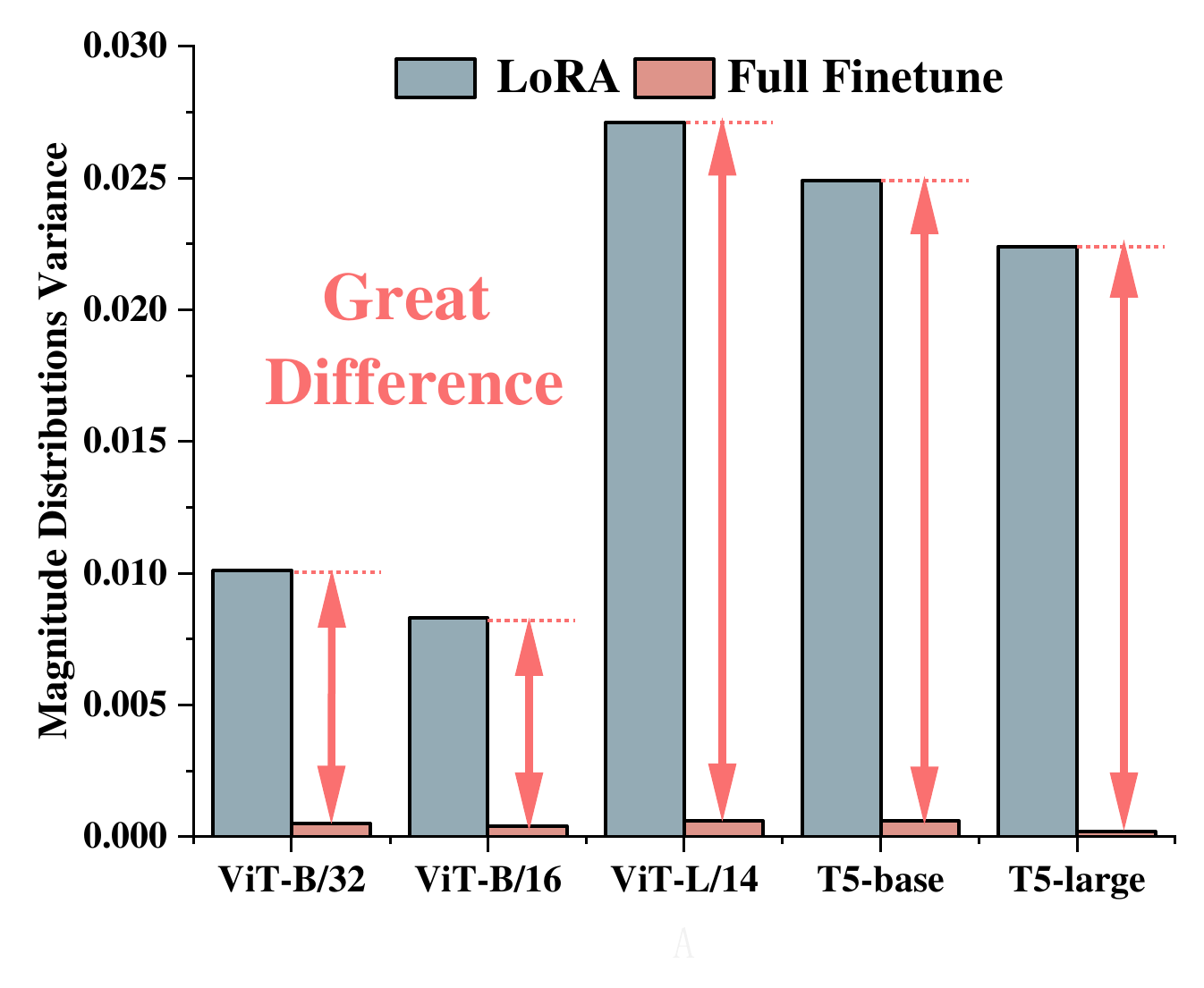}
		\end{minipage}
        \vspace{-0.3cm}
	}%
    \subfigure[]{
		\begin{minipage}[t]{0.32\linewidth}
			\centering
	\includegraphics[width=1\textwidth]{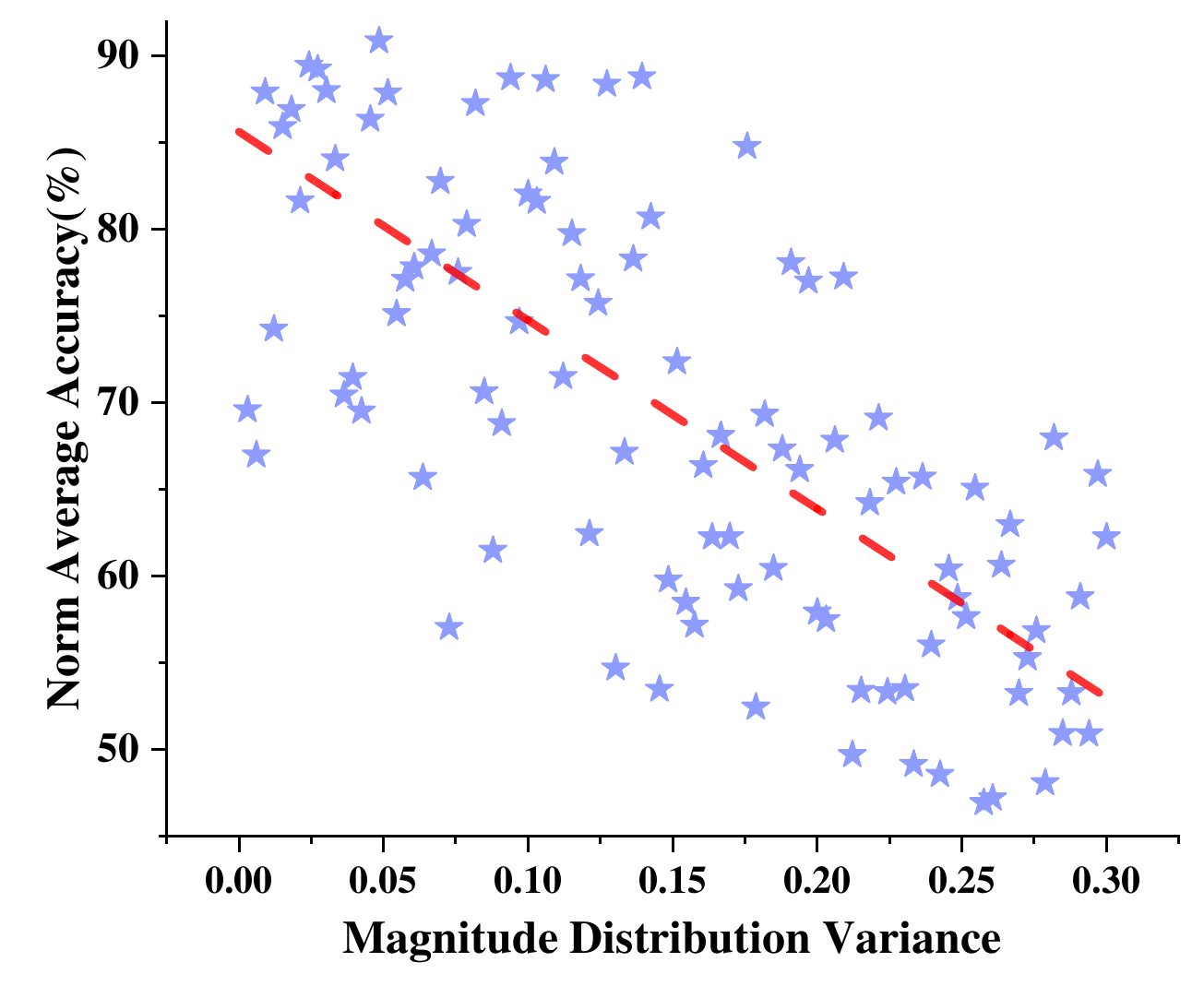}
    
		\end{minipage}
        \vspace{-0.3cm}
	}%
     \vspace{-0.2cm}
	\caption{Key Observations on LoRA Merging. (a) Existing methods work well for full fine-tuning but fail on LoRA. (b) LoRA shows larger parameter discrepancies across tasks than full fine-tuning. The Magnitude Distribution Variance is calculated as discussed in Appendix~\ref{appendix:Computational}.  (c) A greater parameter discrepancy between models correlates with worse merging performance.}
	\label{fig:diff}
    \vspace{-0.4cm}
\end{figure*}

We begin by analyzing why merging strategies designed for full fine-tuning perform poorly on LoRA as shown in Fig.~\ref{fig:diff}(a). We find that LoRA modules trained on different tasks show larger parameter distribution ranges compared to full fine-tuning as shown in Fig.~\ref{fig:diff}(b), and demonstrate in Fig.~\ref{fig:diff}(c) that the performance degradation is closely related to the distribution discrepancies. We propose that this loss stems from the merging process aims to preserve both the distribution scale (magnitude) and shape (direction) of parameters to maintain input mapping patterns~\cite{extend}. However, in LoRA, modules with larger magnitudes dominate the parameter distribution of merged result, leading to shifts in the range and shape of the merged distribution, which causes partial task information loss and hence performance degradation on certain tasks.

\begin{figure*}  
\centering  
\includegraphics[width=1 \textwidth]{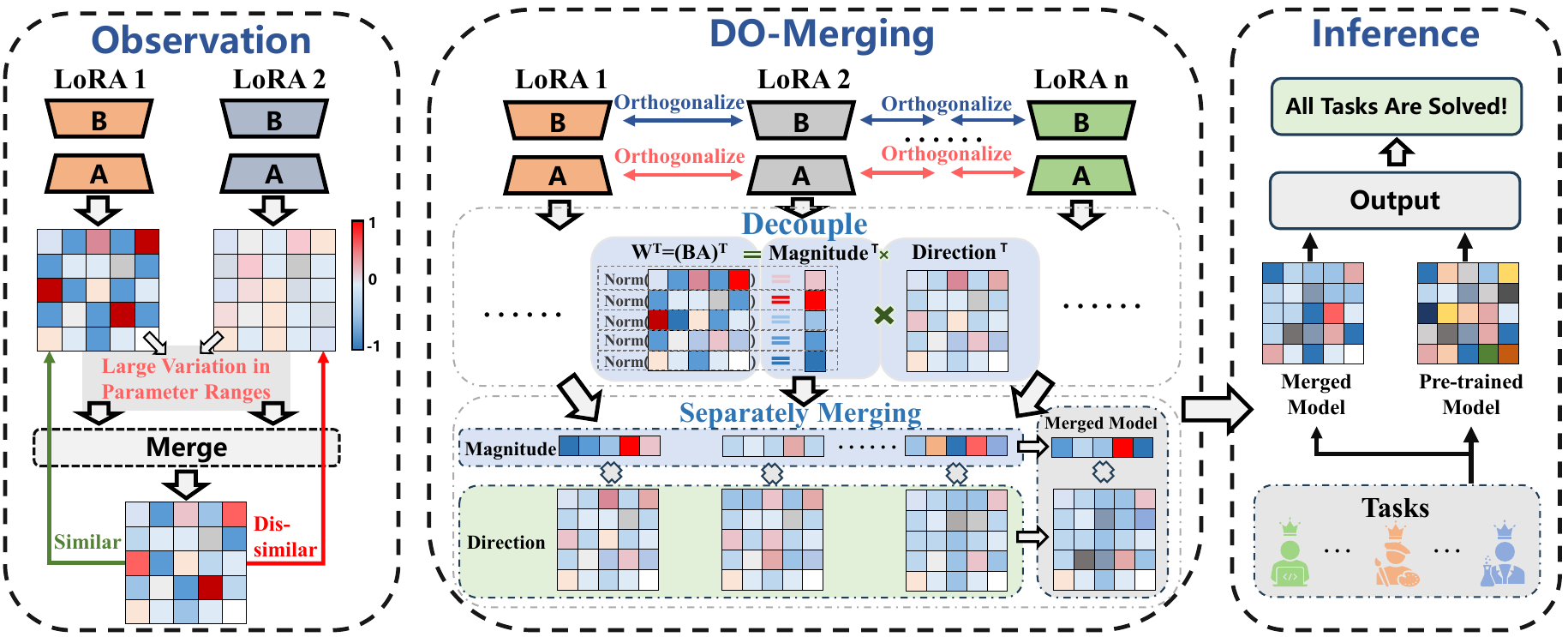} 
\vspace{-0.2cm}
\caption{DO-Merging Framework. Left: Large magnitude differences in LoRA across tasks degrade merging performance. Middle: DO-Merging process—orthogonal perturbation, decoupling magnitude and direction, and separate merging. Right: Single model deployment for multiple tasks.} 
\label{figure:framework.}  
\end{figure*}

To reduce the impact of magnitude on merging, we propose to decouple magnitude and direction, allowing the direction merging to be free from magnitude interference. During direction merging, we further aim to minimize the interference between different tasks in low cost. Then, we introduce \textbf{DO-Merging} (Decouple and Orthogonalize) framework as shown in Fig.~\ref{figure:framework.}.  We first decompose each parameter matrix into a magnitude and direction vector to merge separately, using the normalization coefficient of each column as magnitude and the remaining part as direction. This decoupling ensures that merging is not biased by magnitude differences. As for the magnitude vectors, to retain the scale distributions of each task, we perform average merging. For the direction vectors, we apply data-free, layer-wise gradient descent to generate orthogonal perturbations, minimizing mutual task interference while preserving task-specific information.  Since LoRA merging is performed in the full-rank space as discussed in~\ref{appedix:discussion:seperately merging}, we apply orthogonalization to both LoRA components separately before decoupling in the full-rank matrix for lower cost. This does not affect the conclusion. We then combine the two components to form the final merged model.

We provide theoretical analysis to support DO-Merging’s effectiveness. Experiments across vision, language, and multi-modal tasks confirm its broad applicability. Notably, it achieves over 3\% improvement at minimal cost on various vision tasks, and consistent gains on large language and multi-modal models. And both components can be integrated with existing methods, offering up to nearly 4\% free-lunch improvements across multiple tasks. The main contributions are as follows:
\begin{enumerate}[labelsep = .5em, leftmargin = 0pt, itemindent = 1em]
    \item[$\bullet$] We empirically demonstrate that existing model merging methods for full fine-tuning perform poorly on LoRA. We show that it is closely related to the greater distribution variations exhibited by LoRA compared to full fine-tuning, as supported by both experiments and theory.
    
    \item[$\bullet$] We propose DO-Merging, the first method addressing LoRA merging degradation
    via a decoupled and orthogonal perspective. We decouple parameters into magnitude and direction, merging them separately to mitigate magnitude effects. We further apply a data-free orthogonal constraint to reduce direction interference. Both parts come with theoretical guarantees. 
    
    \item[$\bullet$] We validate the effectiveness of our DO-Merging through comprehensive experiments covering vision, language, and multi-modal tasks. Moreover, both components of DO-Merging can be flexibly combined with current approaches, bringing more than a 3\% performance gain. 
\end{enumerate}

%% file: Sections/Related_Work.tex
\section{Related Work}
\textbf{Model Merging.} Model Merging combines existing models trained on different tasks into a single model that can handle multiple tasks without further training~\cite{ilharco2022editing,jin2022dataless,yang2024model}. The key challenge is resolving task conflicts during merging~\cite{yadav2024ties,qi2024less,hm3}. Existing methods for fully fine-tuned models fall into three categories: automatic computation of merging coefficients~\cite{yangadamerging,jin2022dataless,matena2022merging}, parameter-based conflict reduction~\cite{ilharco2022editing,yangadamerging,du2024parameter,tsvm,qi2024less,gargiulo2024task}, or storing additional knowledge to mitigate interference~\cite{huang2024emr,lu2024twin}. However, these approaches either require extensive data or computation, rely on task-specific designs, or demand major architectural changes, limiting their use by general users. For more details, please refer to Appendix~\ref{appendix:baselines}. When applied to LoRA, these methods suffer from performance drops due to large magnitude differences among modules. To address this, we propose DO-Merging.

\textbf{LoRA Merging.} Current LoRA merging methods fall into two main categories: automatic merging coefficient computation~\cite{iteris,lora.rar,lorahub,lorasoups,copa,Lora-composer} and parameter adjustment to reduce task conflicts~\cite{lego,knots}. The former often requires data or heavy computation, limiting its accessibility for general users. The latter relies on task-specific constraints and custom designs, reducing usability. Additionally, model swarm-style methods require data and suffer from high inference costs, making them impractical for general users.  Details can be found in Appendix~\ref{appendix:baselines}. We find that the key challenge in LoRA merging stems from large magnitude differences among LoRAs. To address this, we propose DO-Merging. 

\textbf{Weight Decouple.} Weight Decouple in deep learning is typically used during training, such as applying distinct objectives for each component~\cite{dora,dota,zhang2025lori}. Decoupling with fixed weights is often used to compute merging coefficients~\cite{extend,weightdisentanglement}. Our work is the first to apply decoupling and orthogonalization in a no-retraining setting for reducing task conflicts during merging.

%% file: Sections/Method.tex
\section{Methodology}\label{sec:method}
\subsection{Preliminary}\label{sec:method:preliminary}
Given a base model $\theta_{pre}$, let $\{\Delta_i\}_{i=1}^n$ denote the LoRA modules for $n$ tasks, each containing $K$ layers. The $k$-th layer of $\Delta_i$ is denoted as a matrix $W_i^k=B_i^kA^k_i$, where $B_i^k$ and $A_i^k$ are the corresponding LoRA parameters. We aim to construct a merged module $\Delta=\lambda \sum_{i=1}^n g(\Delta_i)$ such that the model $\theta_{pre}+\Delta$ can solve all $n$ tasks simultaneously, where $g$ denotes a transformation function.

\subsection{Motivation}\label{sec:method:overview}
As shown in Fig.~\ref{fig:diff}(b), LoRA modules exhibit larger magnitude differences during training than full fine-tuning, and these differences are closely related to the drop in merging performance. Based on this observation, we propose that the merging process should preserve both the distribution shape (direction) and magnitude of the parameters, in order to maintain approximately the same input mapping. However, when the parameter distributions of the models to be merged vary significantly in scale, the directions of the modules with larger ranges tend to dominate, leading to loss of directional information from other modules and thus performance degradation. We provide theoretical support for the relationship between magnitude differences and merging performance. Proof is in Appendix~\ref{appendix:proof3.1}.

\begin{assumption}\label{assump1}
Consider two matrices $W_1, W_2 \in \mathbb{R}^{m \times n}$, and assume $W_i = \alpha_i \times \overline{W_i}$, where $\alpha_i \in \mathbb{R}^{1 \times n}$ is the magnitude and $\overline{W_i}[:,j] \sim \mathcal{N}(0, 1)$. We assume that the merged matrix $W$ preserves features when close to the original matrices, with performance negatively correlated with the loss:
\begin{equation}\label{equa:var_acc}
    L = \frac{\|\alpha_1\|_2 + \|\alpha_2\|_2}{\|\alpha_1\|_2} \|W - W_1\|_2 + \frac{\|\alpha_1\|_2 + \|\alpha_2\|_2}{\|\alpha_2\|_2} \|W - W_2\|_2.
\end{equation}
\end{assumption}

\begin{theorem}\label{theo1}
$\mathbb{E}(L)$ achieves its minimum when $||\alpha_1||_2=||\alpha_2||_2$, and is greater than this minimum in both cases $|| \alpha_1||_2>|| \alpha_2||_2$ and $|| \alpha_1||_2<|| \alpha_2||_2$.
\end{theorem}

Theorem~\ref{theo1} theoretically supports the observation in Fig.~\ref{fig:diff}(c) that large magnitude differences degrade merging performance.
We propose that since magnitude is the key factor causing performance degradation, we decouple magnitude and direction in the merging process. The benefit of this decoupling is to reduce directional information loss caused by magnitude differences during merging. For the magnitude vectors, to preserve the magnitude distribution across models, we merge different magnitude vectors, as detailed in Sec.~\ref{sec:method:decouple}. To reduce merging loss for direction vectors, we propose an efficient, data-free orthogonalization method, also described in Sec.~\ref{sec:method:ortho}. Combining these two components yields the final merged model in Sec.~\ref{sec:method:framework}. Since LoRA involves two low-rank matrix components, we perform orthogonalization on the two low-rank matrices first for reducing cost, and then decouple the full-rank matrices. Although the order is changed, conclusions remain unaffected.

\subsection{Decouple}\label{sec:method:decouple}
First, we introduce a decoupling method to isolate magnitude differences into magnitude vectors, leaving direction vectors with consistent magnitudes. We extract column normalization coefficients as magnitude vectors. For a task vector layer $W=BA \in \mathbb{R}^{m\times n}$, the decoupling process is:
\begin{equation} \label{equa:decouple}
    W = \alpha  \overline{W}, \quad \alpha_{[1,j]} = \text{norm}(W_{[:,j]}).
\end{equation}
Here, $\alpha \in \mathbb{R}^{1\times n}$ represents the magnitude vector, and $\overline{W}$ represents the direction vector. The norm function denotes column normalization of matrix.

Column normalization is chosen as the magnitude vector because for a neural network layer with pre-trained parameters $W_{pre}$ and task-specific parameters $W_t$, the input $x$ and output $x^{out}$ satisfy:
\begin{equation}\label{equa:why_column}
     x^{out} =x (W_{pre}+W_t)=xW_{pre}+xW_t=xW_{pre}+\omega,
\end{equation}
where $\omega_{[i,j]}= \sum_k x_{[i,k]} (W_t)_{[k:j]} $. This indicates that each column of the task vector contributes jointly to the output. Thus, aligning column magnitudes across different task vectors helps preserve their individual output characteristics.

For magnitude vectors, to retain amplitude features from different models after merging, we perform average merging on the magnitude vectors. Assuming we have $n$ task vectors, we define the merged magnitude vector at each layer as: $\alpha_{merge}=\sum_{i=1}^n \alpha_i$. Without optimizing direction vectors, we provide theoretical guarantees by the following theorem that our proposed decoupling method improves merging performance. Detailed proof can be found in Appendix~\ref{appendix:proof3.2}.

\begin{theorem}\label{theorem:3.2}
Let the merging parameters be $W_1$ and $W_2$. The non-decoupled merged model is $W^1 = \lambda_1 \sum_{i=1}^2 \alpha_i \overline{W_i},$ and the decoupled one is $W^2 = \lambda_2\sum_{j=1}^2  \alpha_j \sum_{i=1}^2 \overline{W_i}$. When $\|\alpha_1\|_2 \neq \|\alpha_2\|_2$, the expected loss as Eq.~\ref{equa:var_acc} satisfies $\mathbb{E}(L_2) < \mathbb{E}(L_1)$, which implies that $W^2$ outperforms $W^1$.
\end{theorem}

\subsection{Orthogonalize}\label{sec:method:ortho}

\begin{figure*}[t]
	\centering
	\subfigure[]{
		\begin{minipage}[t]{0.48\linewidth}
			\centering
\includegraphics[width=1\textwidth]{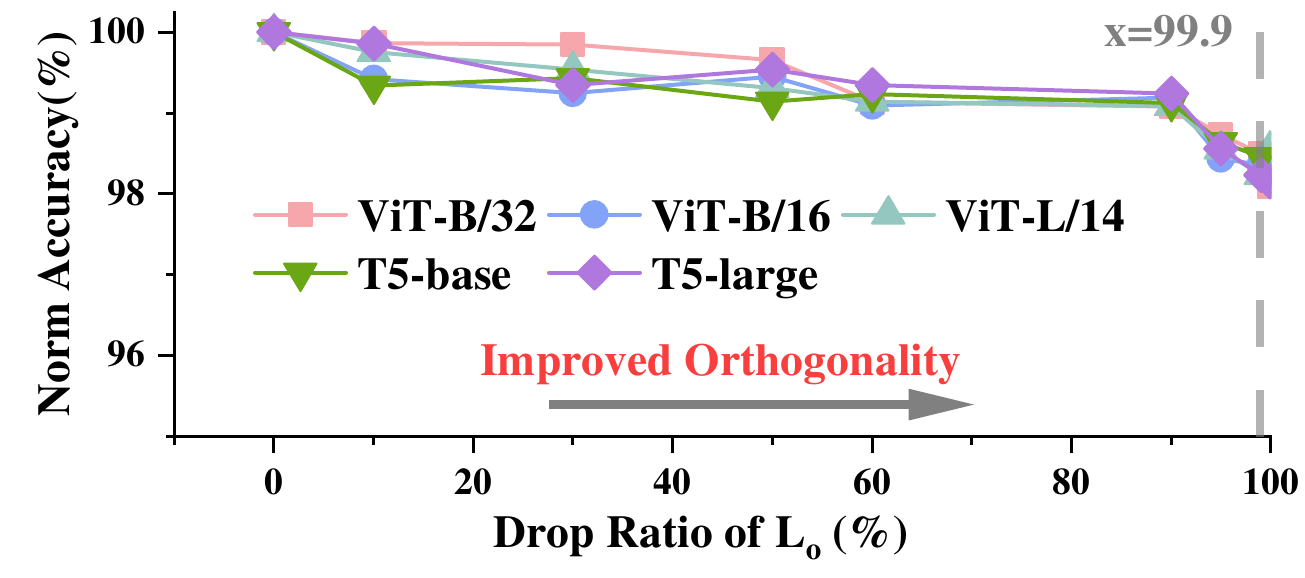} 
		\end{minipage}
        \vspace{-0.3cm}
	}%
    \subfigure[]{
		\begin{minipage}[t]{0.48\linewidth}
			\centering
	\includegraphics[width=1\textwidth]{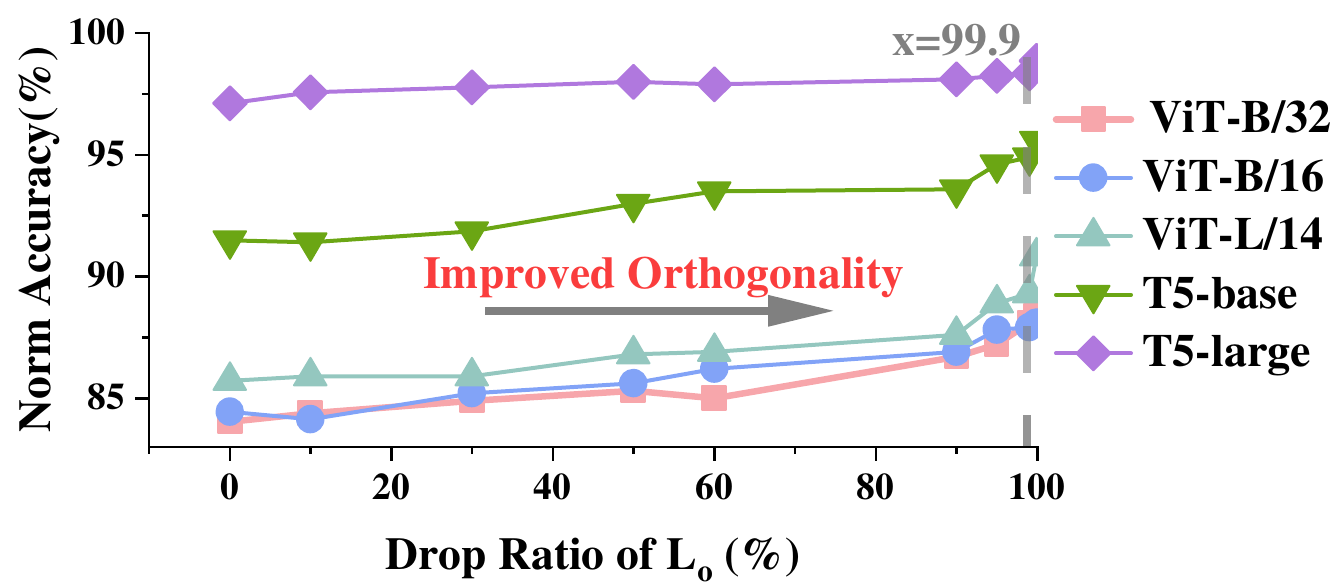}
    
		\end{minipage}
        \vspace{-0.3cm}
	}%
     \vspace{-0.2cm}
	\caption{Key observations on orthogonalization. (a) Average Norm Performance change of task models during orthogonal gradient descent. Performance remains stable. (b) As orthogonality increases, the average merged norm accuracy also improves.} \label{fig:ortho_acc}
	\label{fig:ortho}
    \vspace{-0.4cm}
\end{figure*}

Next, we introduce the optimization of direction vectors. For different task vectors, when two task vectors are nearly orthogonal, it indicates that the corresponding features in the parameter space are also nearly orthogonal, resulting in minimal task interference during merging~\cite{tsvm}. Previous work has shown that imposing orthogonality constraints between LoRA modules during fine-tuning is beneficial for merging~\cite{zhang2025lori}, but publicly available weights often do not meet this requirement. Additionally, ensuring orthogonality using data requires significant amounts of data and substantial forward and backward propagation costs. We aim to achieve similar results using a data-free approach.

Given the high redundancy in fine-tuned parameters, minor adjustments typically do not affect task performance. We therefore propose to apply orthogonality constraints directly to the parameters of existing LoRA modules without data. As illustrated in Fig.~\ref{fig:ortho_acc}(a), this optimization has little negative impact on the performance of individual tasks. Nevertheless, Fig.~\ref{fig:ortho_acc}(b) shows that such orthogonality without considering data factors can still significantly improve the effectiveness of model merging on various models. Therefore, we propose applying data-free and layer-wise parameter-level orthogonality constraints. Thus, for each layer of task vectors, we construct a loss function as follows:
\begin{equation} \label{equa: ortho}
    L=\sum_i \sum_j (W_i+\delta_i)^T (W_j+\delta_j) +\sum_i ||\delta_i||_2=L_o+L_r.
\end{equation}
Since this optimization is layer-wise, the cost is minimal. For LoRA, we apply orthogonality separately on $A$ and $B$, which is equivalent to orthogonality on the product matrix, further reducing cost. We provide theoretical guarantees for the performance improvement due to this orthogonalization by the following theorem. Detailed proof can be found in Appendix~\ref{appendix:proof3.3}.

\begin{theorem}\label{theorem:3.3}
As $||\delta_i||_2 \to 0$, smaller values of  $||W_i^T W_j||$ lead to less conflict during merging.
\end{theorem}
\input{Codes/alg1}

\subsection{Workflow}\label{sec:method:framework}
So far, we have obtained the merged results for both the direction vectors and the magnitude vectors. For each merged weight, we can compute it using the following formula:
\begin{equation}\label{equa:framework}
    W_{out} = W_{pre}+\lambda(\sum_{i=1}^n \alpha_i)(\sum_{j=1}^n \overline{W_j}).
\end{equation}
We also provide the pseudo-code for the overall procedure, as outlined in Alg.~\ref{alg:algorithm1}. We first apply orthogonality constraints separately to the two low-rank matrices of LoRA (lines 2–3). Then, we can obtain orthogonal full-rank matrices. After decomposing magnitude and direction from these matrices, we merge them separately to obtain the final result (lines 5–10).

%% file: Codes/alg1.tex
\begin{wrapfigure}[14]{r}{0.5\textwidth}
  \vspace{-25pt}
  \begin{minipage}{0.48\textwidth}
    \begin{algorithm}[H]
      \caption{Workflow of DO-Merging} 
      \label{alg:algorithm1}
      \begin{algorithmic}[1]
        \REQUIRE Fine-tuned LoRAs $\{B_i,A_i\}_{i=1}^n$ for each layer, pre-trained model $W_{pre}$.
        \STATE \textbf{Step1: Orthogonalize; } 
        \STATE $\{\hat A_i\}_{i=1}^n=Ortho\{ A_i\}_{i=1}^n$ as Eq.~\ref{equa: ortho}.
        \STATE $\{\hat B_i\}_{i=1}^n=Ortho\{ B_i\}_{i=1}^n$ as Eq.~\ref{equa: ortho}.
        \STATE \textbf{Step2: Decouple; }
        \STATE $\{W_i\}_{i=1}^n=\{\hat B_i \hat A_i\}_{i=1}^n$
        \FOR{$i \in [1,n]$}
        \STATE Get $\overline{W_i}, \alpha_i$ as Eq.~\ref{equa:decouple}.
        \ENDFOR
        \STATE \textbf{Step3: Merge; }
        \STATE $W_{out}=W_{pre}+\lambda \sum_{i=1}^n \alpha_i \sum_{j=1}^n \overline{W_j}$
        \ENSURE  Merged model $W_{out}$.
      \end{algorithmic}
    \end{algorithm}
  \end{minipage}
  \vspace{-0pt}
\end{wrapfigure}

%% file: Sections/Experiments.tex
\section{Experiment}
In this part, we conduct comprehensive experimental analyses across various tasks. From Sec.~\ref{sec:exp:vision} to Sec.~\ref{sec:exp:multi}, we perform experiments on multiple tasks using vision models, medium language models, large language models, and multi-modal models, respectively. In Sec.~\ref{sec:exp:ablation}, we present ablation studies about our DO-Merging, and in Sec.~\ref{sec:exp:discussion}, we provide an in-depth discussion of our methods.

\subsection{Vision Models}\label{sec:exp:vision}
\input{Tables/ViT-B-32}
\input{Tables/ViT_16_14_avg}
In this part, we conduct experiments on Vision Transformers (ViT)~\cite{dosovitskiyimage} of different scales. We select ViT-B/32, ViT-B/16, and ViT-L/14, and perform LoRA fine-tuning on each model across eight visual classification tasks, followed by merging the fine-tuned result to evaluate multi-task performance. The detailed fine-tuning and merging configurations are provided in Appendix~\ref{appendix:datasets}. The vision tasks used are SUN397, Cars, RESISC45, EuroSAT, SVHN, GTSRB, MNIST, and DTD~\cite{ilharco2022editing}. The experimental results can be found in Tab~\ref{tab:exp:vision:vit_b_32} and Tab~\ref{tab:exp:vision:vit_16_14_avg}. Additional results are provided in Appendix~\ref{appendix:additional}.

By analyzing the results, we observe that our method demonstrates significant advantages over existing approaches in the merging experiments across all three ViT scales. The average performance improvement across these three ViTs is approximately 2\%. Notably, our method shows substantial gains on certain tasks with all three models, indicating that task-specific parameter features of these tasks suffer from considerable task interference with model merging methods without decoupling and orthogonality. Then, with our DO-Merging, such task interference is reduced, leading to greatly improved performance of the merged model on various target tasks. This strongly demonstrates the effectiveness of our DO-Merging approach.

\subsection{Language Models}\label{sec:exp:language}
\input{Tables/T5_base}
\input{Tables/T5_large}
\input{Tables/LLM}
In this section, we conduct experiments on language models of various scales to demonstrate the generalization capability of our proposed DO-Merging. For medium-sized models, we selected T5-base and T5-large~\cite{raffel2020exploring} as base models and use eight discriminative language tasks as target tasks. These tasks include COLA, MNLI, MRPC, QNLI, QQP, RTE, SST2, and STSB~\cite{tang2024fusionbench}. Detailed fine-tuning information can be found in Appendix~\ref{appendix:datasets}. The detailed experimental results are shown in Tab.~\ref{tab:exp:language:t5_base} and Tab.~\ref{tab:exp:language:t5_large}. It can be observed that for both T5-base and T5-large, our DO-Merging method shows an improvement of over 1\%. This aligns with the analysis of our method’s advantages, indicating that it generalizes well to language models and confirming its strong adaptability.

Additionally, we perform experiments on large-scale language models. We select LLaMa-3-8B~\cite{grattafiori2024llama} and Qwen-14B as base models. For LLaMa-3-8B, we use six natural language understanding tasks as target tasks~\cite{knots}, while for Qwen-14B~\cite{bai2023qwen}, we chose MMLU~\cite{hendrycksmeasuring}, TruthfulQA~\cite{lin2022truthfulqa}, and BBQ~\cite{parrish2022bbq} as target tasks. Detailed fine-tuning information is provided in Appendix~\ref{appendix:datasets}. The experimental results are presented in Tab.~\ref{tab:exp:language:llm}. In the experiments on LLaMa3-8B, our proposed method shows an improvement of nearly 2\% on average, while existing merging methods do not offer significant gains over the basic baseline, Task Arithmetic. Similarly, in the Qwen-14B experiments, DO-Merging achieves more than a 1\% average improvement. This aligns with our analysis that addressing the performance loss due to magnitude differences in LoRA merging is crucial. This performance enhancement further confirms the superior performance and generalization ability of DO-Merging.

\subsection{Multi-modal Models}\label{sec:exp:multi}
\input{Tables/MLLM}
To further evaluate the generalization ability of DO-Merging, we conduct experiments on multi-modal tasks. We use Qwen2-VL~\cite{Qwen2-VL} as the base model and fine-tune then test it on five multi-modal tasks: POPE~\cite{pope}, MMStar~\cite{mmstar}, MMBench~\cite{mmbench}, RealWorldQA~\cite{xai2025grok3}, and MathVista~\cite{mathvista}. Detailed weight information can be found in the Appendix. Among these tasks, POPE focuses on detecting model hallucinations, while MMStar and MMBench evaluate different aspects of multi-modal capabilities from diverse metrics. RealWorldQA requires the model to understand real-world scenarios, and MathVista tests its mathematical reasoning ability. These tasks vary significantly in terms of required skills, making them relatively challenging for model merging. As shown in Tab.~\ref{tab:multi-modal}, our DO-Merging method achieves the best performance across all tasks, and even outperforms the fine-tuned model on MMStar. This demonstrates that our method can be effectively applied to multi-modal tasks. Moreover, by selecting tasks with low correlation, we verify that DO-Merging still improves merging performance, further confirming its strong generalization capability and superior performance.

\input{Tables/Ablation}
\subsection{Ablation Study}\label{sec:exp:ablation}
In this section, we conduct ablation studies on our method. 

\textbf{Decouple and Orthogonalize.} We present ablation results for combining the two components we proposed in DO-Merging. We perform experiments on ViT-B/32, ViT-B/16, and ViT-L/14, using eight visual classification tasks as target tasks. The detailed results are shown in Tab.~\ref{tab:ablation_all}. Compared to methods that include neither component, adding only the orthogonal part leads to an average improvement of over 2\%, while adding only the decoupling part results in an improvement of over 1\%. This holds true across all three ViT models. When both components are used together, the performance gain is even larger. This strongly demonstrates the effectiveness of both components in DO-Merging, which aligns with our previous theoretical and experimental analysis.

\textbf{Magnitude Extraction.} In Tab.~\ref{tab:decouple}, we show results for using different parts of the parameter matrix as the magnitude vector in decoupling. \textit{Matrix Norm} means using the normalization coefficient of the entire parameter matrix as the magnitude vector, while \textit{Row Norm} and \textit{Col Norm} refer to using the row-wise and column-wise normalization coefficients, respectively. From the comparison, we observe that using the matrix norm does not perform well, likely due to its coarse granularity, which fails to preserve fine-grained features during merging. Consistent with our analysis, using the column norm achieves better performance than using the row norm. This is because, in a parameter matrix, each column is responsible for one output dimension, and during merging, we aim to keep the merged output as close as possible to the original models' outputs. Therefore, it is reasonable to use the column norm as the magnitude vector in our decoupling design.

\subsection{Discussions}\label{sec:exp:discussion}
\begin{figure*}[t]
	\centering
	\subfigure[Flexible Combination]{
		\begin{minipage}[t]{0.33\linewidth}
			\centering
		\includegraphics[width=1\textwidth]{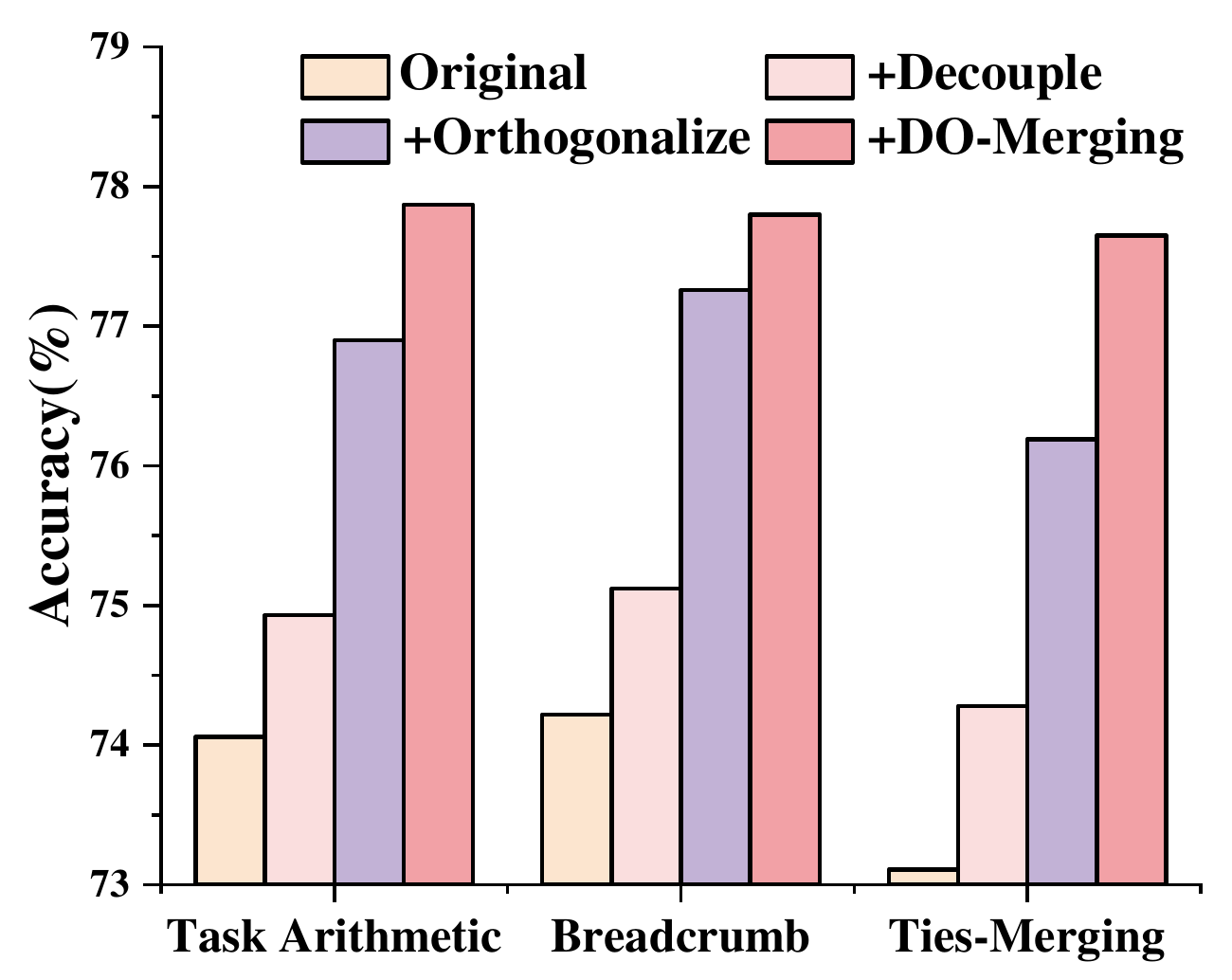}
		\end{minipage}
	}%
	\subfigure[Orthogonal Effect]{
		\begin{minipage}[t]{0.33\linewidth}
			\centering
	\includegraphics[width=1\textwidth]{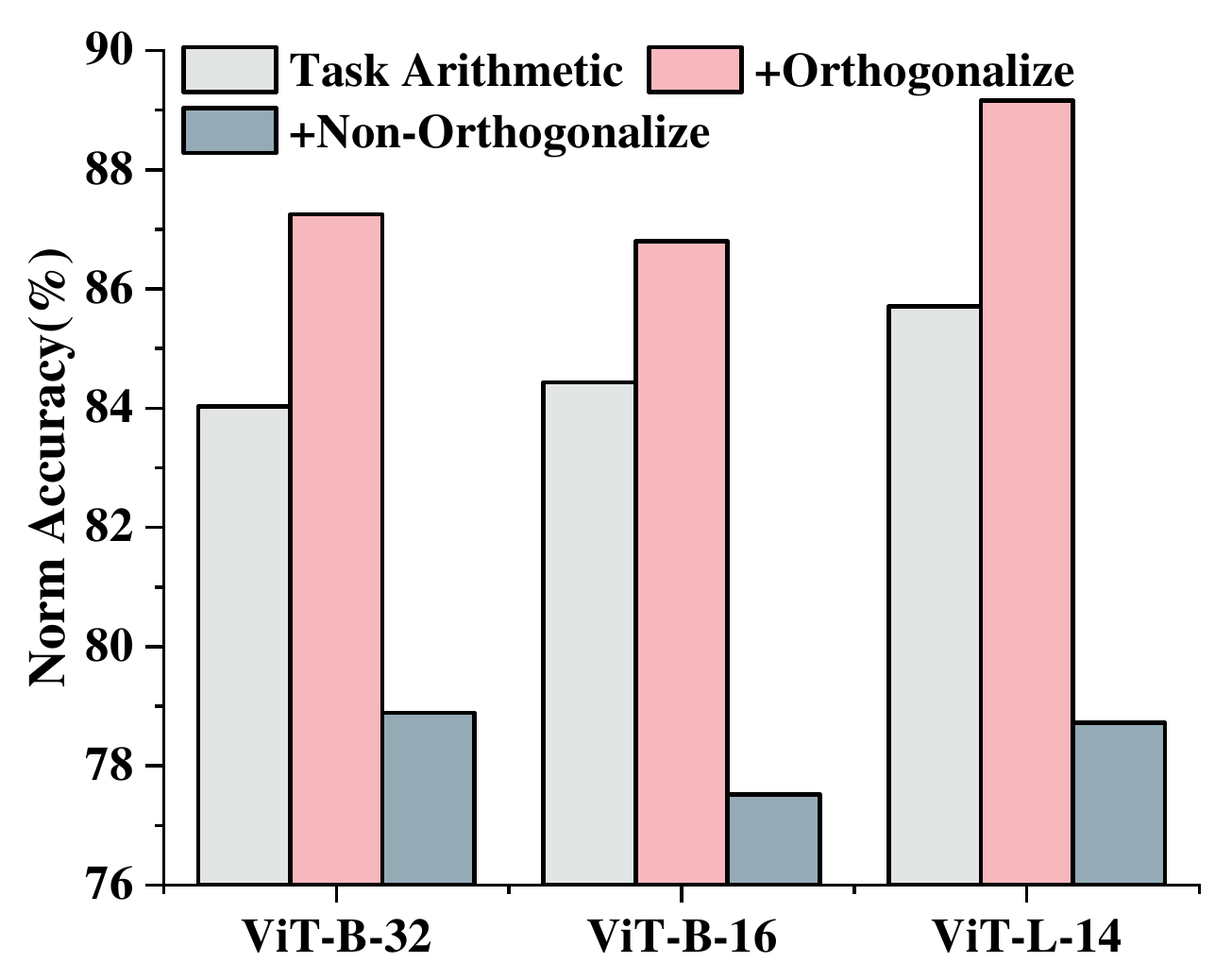}
		\end{minipage}
	}%
    \subfigure[The Impact of LoRA Rank]{
		\begin{minipage}[t]{0.33\linewidth}
			\centering
	\includegraphics[width=1\textwidth]{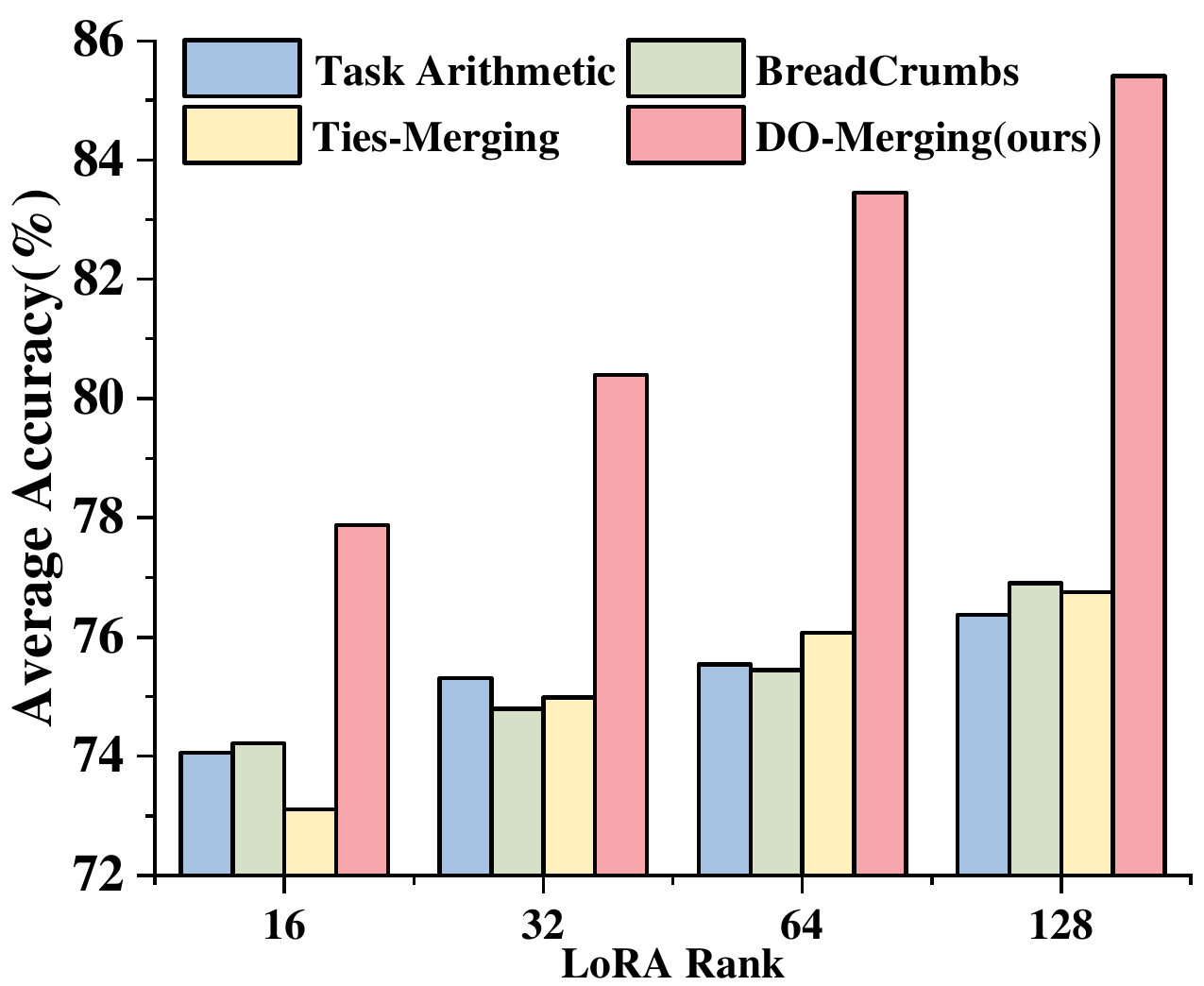}
		\end{minipage}
	}%
     \vspace{-0.2cm}
	\caption{Discussion on Key Properties of DO-Merging. (a). Both components of DO-Merging can be freely combined with other merging methods, and bring near free-lunch improvement. (b). Orthogonality in LoRA is crucial. Performance drops significantly when the direction vectors are made non-orthogonal. (c). Our method performs well across different LoRA ranks.}
	\label{fig:discussions}
    \vspace{-5pt}
\end{figure*}
We provide a broader discussion of our DO-Merging. More discussion is in Appendix~\ref{appendix:discussion}.

\textbf{Flexible Combination.} We discuss how our decoupling and orthogonal methods can be flexibly combined with other model merging approaches to further improve their performance. We integrate our proposed components with two popular existing methods, Breadcrumbs and Ties-Merging, and conduct experiments on ViT-B/32. Detailed results are shown in Fig.~\ref{fig:discussions}(a). It can be observed that whether adding only the decoupling component or only the orthogonal component, our method consistently improves the performance of existing merging techniques. Both components demonstrate strong compatibility with different methods, allowing users to choose based on practical needs or preferred base model merging strategies. This highlights the flexibility of our decoupling and orthogonal approaches, which significantly benefits the applicability.

\textbf{Orthogonal Effect.} Here, we examine whether our proposed orthogonal method plays a key role in improving merging performance. In Fig.~\ref{fig:discussions}(b), we compare the effectiveness of our orthogonal approach by using gradient descent and gradient ascent during the optimization process. We find that when the orthogonality between different LoRA modules is reduced (i.e., increased non-orthogonality), the average accuracy of LoRA merging drops significantly. This strongly indicates that orthogonality is a decisive factor in maintaining performance during model merging. Therefore, our design of enforcing orthogonality among LoRA parameters is well justified.

\begin{wrapfigure}[17]{r}{0.45\textwidth} 
  \vspace{-1.8em}  
  \centering
  \includegraphics[width=0.9\linewidth]{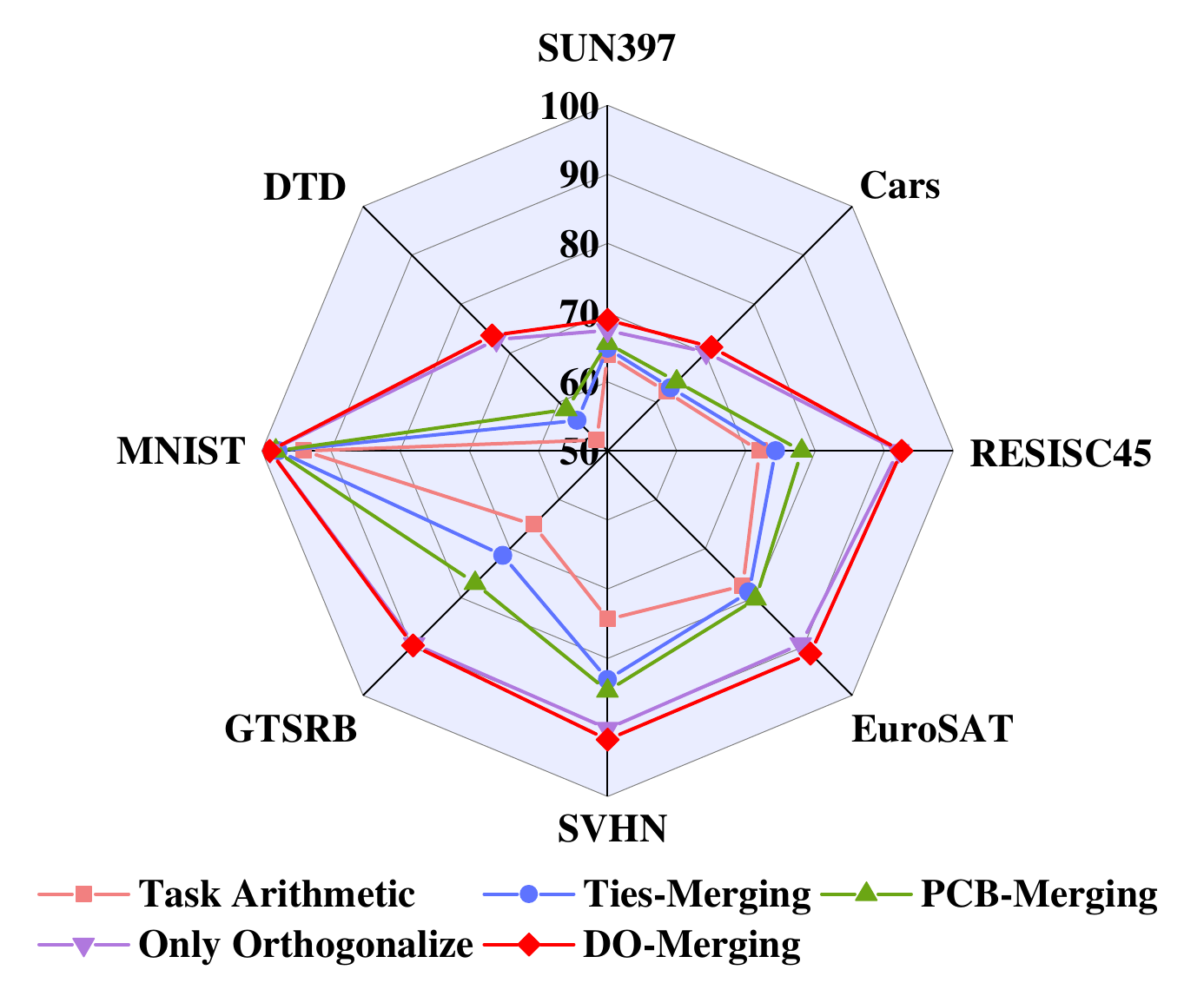}
  \caption{Applying DO-Merging to the merging of fully fine-tuned models is also effective. Experiments are conducted on ViT-B/32.} 
  \label{fig:transfer_fft}
  \vspace{-1em}
\end{wrapfigure}

\textbf{The Impact of LoRA Rank.} Most experiments in the previous sections used a LoRA rank of 16, a common configuration. Here, we evaluate our method's effectiveness across various ranks. Using ViT-B/32 as the base model, we tested DO-Merging on eight visual tasks with LoRA ranks of 16, 32, 64, and 128. The results in Fig.~\ref{fig:discussions}(c) show that our method consistently outperforms popular model merging techniques across all ranks. Additionally, performance improves with increasing rank, demonstrating the robustness and superiority of our approach.

\textbf{Transfer to Full-Finetune Merging.} In this part, we evaluate its effectiveness of DO-Merging when transferred to full-finetune merging. We use ViT-B/32 as the base model and test on eight vision tasks. The results are shown in Fig.~\ref{fig:discussions}(c). It can be observed that orthogonalization remains highly effective. And the gain from decoupling is less significant compared to LoRA merging, which aligns with our observation that full fine-tuning leads to smaller magnitude differences across models. Nevertheless, our method still performs well under full fine-tuning, demonstrating its generalization ability.

%% file: Tables/ViT-B-32.tex
\begin{table}[t]
    \centering
    \caption{Multi-task performance when merging ViT-B/32 on eight vision tasks.}\label{tab:exp:vision:vit_b_32}
    \vspace{3pt}
    \scalebox{0.82}{
    \begin{tabular}{l|cccccccc|c}
    \toprule
        Method & SUN397 & Cars & RESISC45 & EuroSAT & SVHN & GTSRB & MNIST & DTD & Avg. \\ \midrule
        Pre-trained & 62.30 & 59.70 & 60.70 & 45.50 & 31.40 & 32.60 & 48.50 & 43.80 & 48.06 \\ 
        Finetune & 71.81 & 71.23 & 94.33 & 98.85 & 97.19 & 98.62 & 99.63 & 73.40 & 88.13 \\ \midrule
        Task Arithmetic~\cite{ilharco2022editing} & 63.38 & 60.27 & 75.46 & 87.70 & 84.73 & 70.74 & 96.64 & 53.55 & 74.06 \\ 
        Ties Merging~\cite{yadav2024ties} & 63.81 & 53.95 & 73.56 & 86.00 & 88.27 & 71.15 & 97.81 & 50.32 & 73.11 \\ 
        Breadcrumbs~\cite{davari2023model} & 64.02 & 57.87 & 74.12 & 88.73 & 84.24 & 74.28 & 97.29 & 53.23 & 74.22 \\ 
        AdaMerging~\cite{yangadamerging} & 62.59 & 57.89 & 75.94 & \textbf{90.63} & 78.93 & 88.82 & 96.92 & 52.07 & 75.47 \\ 
        PCB-Merging~\cite{du2024parameter} & 63.74 & 58.64 & 73.58 & 88.12 & 85.65 & 75.12 & 97.45 & 54.07 & 74.54 \\ 
        KNOTS~\cite{knots} & 64.08 & 59.26 & 76.11 & 86.93 & 86.94 & 75.54 & 96.85 & 57.46 & 75.39 \\ 
        TSVM~\cite{tsvm} & \textbf{64.81} & 59.31 & \textbf{78.73} & 89.63 & 85.43 & 76.63 & 97.02 & 57.93 & 76.19 \\ 
        CoPA-Merging~\cite{copa} & 63.40 & 59.56 & 76.68 & 88.93 & 85.54 & 77.00 & 97.14 & 56.56 & 75.60 \\ \midrule
        DO-Merging(ours) & 64.48 & \textbf{61.01} & 77.30 & 88.30 & \textbf{89.05} & \textbf{83.31} & \textbf{98.03} & \textbf{61.53} & \textbf{77.88} \\ \bottomrule
    \end{tabular}
    }
    \vspace{-10pt}
\end{table}

%% file: Tables/ViT_16_14_avg.tex
\begin{wraptable}{r}{7cm}
    \centering
    \vspace{-12pt}
    \caption{Average multi-task performance when merging ViT-B/16 and ViT-L-14 on eight tasks.}\label{tab:exp:vision:vit_16_14_avg}
    \scalebox{0.80}{
    \begin{tabular}{lcc}
    \toprule
         \textbf{Method} & \textbf{ViT/B-16} & \textbf{ViT/L-14} \\ \midrule
        Pre-trained & 55.20 & 64.89 \\
        Finetune & 89.94 & 93.11 \\ \midrule
        Task Arithmetic~\cite{ilharco2022editing} & 75.94 & 79.80 \\ 
        Ties Merging~\cite{yadav2024ties} & 76.18({\color{blue}$\uparrow 0.24\%$}) & 79.28({\color{red}$\downarrow 0.52\%$}) \\ 
        Breadcrumbs~\cite{davari2023model} & 76.00({\color{blue}$\uparrow 0.06\%$}) & 79.60({\color{red}$\downarrow 0.20\%$}) \\ 
        AdaMerging~\cite{yangadamerging} & 77.93({\color{blue}$\uparrow 1.99\%$}) & 82.83({\color{blue}$\uparrow 3.03\%$}) \\ 
        PCB-Merging~\cite{du2024parameter} & 77.19({\color{blue}$\uparrow 1.25\%$}) & 80.79({\color{blue}$\uparrow 0.99\%$}) \\ 
        KNOTS~\cite{knots} & 76.51({\color{blue}$\uparrow 0.57\%$}) & 80.95({\color{blue}$\uparrow 1.05\%$}) \\ 
        TSVM~\cite{tsvm} & 77.19({\color{blue}$\uparrow 1.25\%$}) & 80.86({\color{blue}$\uparrow 1.06\%$}) \\ 
        CoPA-Merging~\cite{copa} & 76.79({\color{blue}$\uparrow 0.85\%$}) & 81.43({\color{blue}$\uparrow 1.63\%$}) \\ \midrule
        DO-Merging(ours) & \textbf{79.24({\color{blue}$\uparrow 3.30\%$)}} & \textbf{84.58({\color{blue}$\uparrow 4.78\%$)}} \\ \bottomrule
    \end{tabular}
    }
\end{wraptable}

%% file: Tables/T5_base.tex
\begin{table}[t]
    \centering
    \caption{Multi-task performance when merging T5-base on eight language tasks.}\label{tab:exp:language:t5_base}
    \vspace{3pt}
    \scalebox{0.9}{
    \begin{tabular}{l|cccccccc|c}
    \toprule
        Method& COLA & MNLI & MRPC & QNLI & QQP & RTE & SST2 & STSB & Avg. \\  \midrule
        Pre-trained & 69.1 & 56.5 & 76.2 & 88.4 & 82.1 & 80.1 & 91.2 & 62.2 & 75.7 \\ 
        Finetune & 69.1 & 82.7 & 85.5 & 90.9 & 84.0 & 84.5 & 92.9 & 87.4 & 84.6 \\ \midrule
        Task Arithmetic~\cite{ilharco2022editing} & 68.8 & 55.2 & 78.7 & 89.8 & 83.7 & 79.1 & 91.5 & 72.4 & 77.4 \\ 
        Ties Merging~\cite{yadav2024ties} & 68.3 & 56.3 & 79.4 & 89.8 & 83.7 & 79.4 & 91.6 & 71.2 & 77.5 \\ 
        Breadcrumbs~\cite{davari2023model} & 68.3 & 57.3 & 78.5 & 89.8 & 83.6 & 79.5 & 91.8 & 72.4 & 77.6 \\
        PCB-Merging~\cite{du2024parameter} & 68.3 & 57.4 & 79.3 & 89.8 & 83.6 & 79.2 & 91.5 & 72.6 & 77.7 \\ 
        KNOTS~\cite{knots} & 69.4 & 63.6 & 78.5 & 89.8 & 83.6 & 80.2 & 91.5 & 70.6 & 78.4 \\ 
        TSVM~\cite{tsvm} & 69.1 & 67.8 & \textbf{79.9} & 89.7 & 83.6 & \textbf{80.9} & 91.6 & 69.4 & 79.0 \\ 
        CoPA-Merging~\cite{copa} & \textbf{69.4} & 59.0 & 79.2 & 89.9 & 83.6 & 80.1 & 91.4 & \textbf{75.4} & 78.5 \\ \midrule
        DO-Merging(ours) & 69.3 & \textbf{75.8} & 79.7 & \textbf{90.5} & \textbf{84.0} & 80.1 & \textbf{93.1} & 74.4 & \textbf{80.9} \\ \bottomrule
    \end{tabular}
}
\end{table}

%% file: Tables/T5_large.tex
\begin{table}[t]
    \centering
    \caption{Multi-task performance when merging T5-large on eight language tasks.}\label{tab:exp:language:t5_large}
    \vspace{3pt}
    \scalebox{0.9}{
    \begin{tabular}{l|cccccccc|c}
    \toprule
        Method& COLA & MNLI & MRPC & QNLI & QQP & RTE & SST2 & STSB & Avg. \\  \midrule
        Pre-trained & 73.7 & 56.6 & 82.4 & 91.1 & 85.5 & 85.6 & 94.3 & 87.5 & 82.1 \\ 
        Finetune & 80.2 & 88.5 & 89.2 & 94.4 & 87.2 & 91.7 & 95.2 & 90.9 & 89.6 \\ \midrule
        Task Arithmetic~\cite{ilharco2022editing} & 76.9 & 85.4 & 85.3 & \textbf{93.9} & 85.8 & 88.1 & 95.2 & 87.8 & 87.3 \\ 
        Ties Merging~\cite{yadav2024ties} & 77.1 & 85.1 & 86.3 & \textbf{93.9} & 86.0 & 87.7 & 95.1 & 88.0 & 87.4 \\ 
        BreadCrumbs~\cite{davari2023model} & 77.2 & 85.2 & 86.1 & \textbf{93.9} & 86.0 & 87.4 & 94.9 & 88.4 & 87.4 \\ 
        PCB-Merging~\cite{du2024parameter} & 77.3 & 85.4 & 85.8 & \textbf{93.9} & 86.0 & 87.9 & 95.2 & 88.0 & 87.4 \\ 
        KNOTS~\cite{knots} & 76.8 & 87.4 & 86.4 & 93.5 & 86.3 & 87.3 & 95.2 & 88.6 & 87.7 \\ 
        TSVM~\cite{tsvm} & 76.0 & 88.0 & 84.3 & 92.9 & 86.3 & 87.4 & 95.2 & 88.5 & 87.3 \\ 
        CoPA-Merging~\cite{copa} & 76.0 & 88.4 & 86.5 & 93.0 & 86.1 & \textbf{87.7} & 95.3 & 88.4 & 87.7 \\ \midrule
        DO-Merging(ours) & \textbf{78.9} & \textbf{88.5} & \textbf{88.2} & 93.8 & \textbf{86.5} & 87.4 & \textbf{96.1} & \textbf{89.2} & \textbf{88.6} \\ \bottomrule
    \end{tabular}
    }
    \vspace{-8pt}
\end{table}

%% file: Tables/LLM.tex
\begin{table}[!ht]
    \centering
    \caption{The merging performance of LLaMa3-8B and Qwen-14B on the corresponding tasks.}\label{tab:exp:language:llm}
    \vspace{3pt}
    \scalebox{0.75}{
    \begin{tabular}{l|ccccccc|cccc}
    \toprule
        Base Model & \multicolumn{7}{c|}{LLaMa3-8B} & \multicolumn{4}{c}{Qwen-14B} \\ \midrule
        Task & SNLI & MNLI & SICK & QNLI & RTE & SCITAIL & Avg. & MMLU & TruthfulQA & BBQ & Avg. \\ \midrule
        Pre-trained & 41.69 & 34.64 & 55.63 & 51.92 & 42.03 & 60.72 & 47.77 & 69.30 & 51.27 & 80.69 & 67.09 \\ 
        Finetune & 92.49 & 90.30 & 91.58 & 94.48 & 89.85 & 96.51 & 92.53 & 68.35 & 54.34 & 93.53 & 72.07 \\ \midrule
        Task Arithmetic~\cite{ilharco2022editing} & 86.56 & 86.05 & 80.57 & 64.91 & 89.85 & 93.36 & 83.55 & 67.62 & 53.38 & 78.24 & 66.41 \\ 
        Ties Merging~\cite{yadav2024ties} & 86.48 & 86.15 & 80.43 & 65.04 & 89.85 & 93.75 & 83.61 & 68.27 & 50.01 & 84.10 & 67.46 \\ 
        KNOTS~\cite{knots} & 87.95 & 86.94 & 83.03 & 67.20 & 89.58 & 93.58 & 84.71 & 68.20 & 52.48 & 83.56 & 68.08 \\ 
        TSVM~\cite{tsvm} & 88.82 & 83.96 & 81.79 & 74.87 & 91.30 & 94.02 & 85.79 & 68.10 & 52.24 & 82.42 & 67.58 \\ 
        CoPA-Merging~\cite{copa} & 88.59 & 87.49 & \textbf{84.59} & 67.60 & 89.85 & 93.67 & 85.29 & 68.10 & 52.11 & 83.75 & 67.98 \\ \midrule
        DO-Merging(ours) & \textbf{89.05} & \textbf{88.05} & 83.14 & \textbf{77.13} & \textbf{91.30} & \textbf{94.02} & \textbf{87.11} & \textbf{68.45} & \textbf{53.88} & \textbf{84.99} & \textbf{69.10} \\ \bottomrule
    \end{tabular}
    }
\end{table}

%% file: Tables/MLLM.tex
\begin{table}[t]
    \centering
    \caption{Multi-task performance when merging Qwen2-VL on multi-modal tasks.}\label{tab:multi-modal}
    \vspace{3pt}
    \scalebox{0.9}{
    \begin{tabular}{l|ccccc|c}
    \toprule
        Method & POPE & MMStar & MathVista & RealworldQA & MMBench & Avg. \\ \midrule
        Pre-trained & 85.86 & 59.33 & 57.80 & 69.80 & 80.32 & 70.62 \\ 
        Finetune & 97.75 & 63.60 & 64.00 & 73.46 & 97.50 & 79.26 \\ \midrule
        Task Arithmetic~\cite{ilharco2022editing} & 91.27 & 64.60 & 61.50 & 69.67 & 90.98 & 75.60 \\ 
        Ties Merging~\cite{yadav2024ties} & 91.27 & 64.80 & 61.40 & 70.42 & 91.13 & 75.80({\color{blue}$\uparrow 0.20\%$}) \\ 
        Breadcrumbs~\cite{davari2023model} & 92.83 & 66.40 & 60.69 & 69.67 & 90.98 & 76.11({\color{blue}$\uparrow 0.51\%$}) \\ 
        KNOTS~\cite{knots} & 92.42 & 65.64 & 61.55 & 70.46 & 92.58 & 76.53({\color{blue}$\uparrow 0.93\%$}) \\ 
        TSVM~\cite{tsvm} & 92.74 & 65.26 & 61.60 & 70.21 & 93.49 & 76.66({\color{blue}$\uparrow 1.06\%$}) \\ 
        CoPA-Merging~\cite{copa} & 93.32 & 65.33 & 62.50 & 70.31 & 93.45 & 76.98({\color{blue}$\uparrow 1.38\%$}) \\ \midrule
        DO-Merging(ours) & \textbf{94.74} & \textbf{66.80} & \textbf{62.80} & \textbf{72.13} & \textbf{96.30} & \textbf{78.50}({\color{blue}$\uparrow$ 2.90\%}) \\ \bottomrule
    \end{tabular}
    }
    \vspace{-10pt}
\end{table}

%% file: Tables/Ablation.tex
\begin{minipage}[t]{0.49\linewidth}
\centering
\setlength{\tabcolsep}{0.7 mm}
 \captionof{table}{Ablation study on the proposed decoupling and orthogonal components. \ding{55} indicates that the module is not included, while \ding{51} indicates that the module is included.}\label{tab:ablation_all}
    \vskip -0.03in
    \scalebox{0.88}{
        \begin{tabular}{ccccc}
    \toprule
        Decouple & Orthogonalize & ViT-B/32 & ViT-B/16 & ViT-L/14 \\ \midrule
        \ding{55} & \ding{55} & 74.06 & 75.94 & 79.80 \\ 
        \ding{55} & \ding{51} & 76.90 & 78.07 & 83.02 \\ 
        \ding{51} & \ding{55} & 74.93 & 77.05 & 83.20 \\ 
        \ding{51} & \ding{51} & \textbf{77.87} & \textbf{79.24} & \textbf{84.58} \\ \bottomrule
    \end{tabular}
    }
\end{minipage}
\hfill
\begin{minipage}[t]{0.49\linewidth}
\centering
\setlength{\tabcolsep}{0.7mm}
\captionof{table}{Ablation study on magnitude vectors: \textit{matrix norm} uses the full parameter matrix’s normalization coefficients, while \textit{row norm} and \textit{col norm} use row-wise and column-wise coefficients.}\label{tab:decouple}
    \vskip -0.03in
    \scalebox{0.9}{
    \begin{tabular}{lccc}
    \toprule
         ~& ViT-B/32 & ViT-B/16 & ViT-L/14 \\ \midrule
        Task Arithmetic & 74.06 & 75.94 & 79.80 \\ 
        Matrix Norm & 76.98 & 76.21 & 77.45 \\ 
        Row Norm & 76.89 & 76.67 & 82.52 \\ 
        Col Norm & \textbf{77.87} & \textbf{79.24} & \textbf{84.58} \\ \bottomrule
    \end{tabular}
    }
\end{minipage}

%% file: Sections/Conclusion.tex
\section{Conclusion}
In this paper, we address the issue that existing model merging methods do not perform well when directly applied to LoRA. We find that this is mainly due to great parameter magnitude differences that commonly arise during LoRA training. To tackle this problem, we propose a LoRA merging method in which we decouple the magnitude and direction of the parameters. This helps reduce the loss of directional information during merging caused by magnitude variations. Furthermore, to reduce task conflicts, we apply data-free orthogonal perturbations to the direction vectors. Based on these ideas, we propose our DO-Merging. This approach does not require access to training data and can be flexibly used in different combinations, making it a promising solution for LoRA merging.

%% file: Sections/Appendix.tex
\vspace{20pt}
\textbf{{\Large Appendix for DO-Merging}}

\section{Notations}\label{appendix:notations}
\input{Tables/Notations}
Tab.~\ref{tab:appendix:notations} presents the main notations used in this paper and their corresponding meanings. It should be noted that, for simplicity in explanation and proof, the layer index $k$ is sometimes omitted when it does not affect the clarity of the discussion or derivation.

\section{Proofs}\label{appendix:Proofs}
In this section, we provide a proof for the theorem presented in the original paper.

\subsection{Theorem 3.1}\label{appendix:proof3.1}
It follows from the Assumption 3.1 that:
\begin{equation}
    W_1=\alpha_1 \overline{W_1}, W_2=\alpha_2 \overline{W_2}
\end{equation}
The simplest merging model can be expressed as:
\begin{equation}
    W=\frac{1}{2}(W_1+W_2)
\end{equation}
Let $||\alpha_1||_2=\lambda||\alpha_2||_2$, Then we have:
\begin{align}
    L&=\frac{||\alpha_1||_2+||\alpha_2||_2}{||\alpha_1||_2}||W- W_1||_2+\frac{||\alpha_1||_2+||\alpha_2||_2}{||\alpha_2||_2}||W- W_2||_2\\
&=\frac{\lambda^2+1}{\lambda^2}||\alpha_2||_2 ||\frac{1}{2}(W_2-W_1)||_2+(\lambda^2+1)||\alpha_2||_2 ||\frac{1}{2}(W_2-W_1)||_2
\end{align}
Because $\overline{W_{1}}_{ij} \sim N(0,1)$,and $\overline{W_{2}}_{ij} \sim N(0,1)$.

Then we have:
\begin{align}\label{appendix:proof_3.1_EL}
\mathbb{E}(L)&=\mathbb{E}\{(\lambda^2+1)[\sum_{i=1}^m \sum_j \frac{1}{\lambda^2}(\frac{1}{4} \alpha_{1j}^2 (\overline{W_1})_{ij}^2+\frac{1}{4} \alpha_{2j}^2 (\overline{W_2})_{ij}^2-\frac{1}{2}\alpha_{1j}\alpha_{2j}(\overline{W_1})_{ij}(\overline{W_2})_{ij})\\
&+(\frac{1}{4} \alpha_{1j}^2 (\overline{W_1})_{ij}^2+\frac{1}{4} \alpha_{2j}^2 (\overline{W_2})_{ij}^2-\frac{1}{2}\alpha_{1j}\alpha_{2j}(\overline{W_1})_{ij}(\overline{W_2})_{ij})]\}\\
&=(\lambda^2+1)[\sum_{i=1}^m \sum_j \frac{1}{4}(1+\frac{1}{\lambda^2}) (\alpha_{1j}^2+\alpha_{2j}^2)]\\
&=n(\lambda^2+1) ||\alpha_2||_2[(\frac{1}{\lambda^2}+1)(\frac{1}{4}\lambda^2+\frac{1}{4})]\\
&=n(\lambda^2+1)||\alpha_2||_2(\frac{1}{4}\lambda^2+\frac{1}{2}+\frac{1}{4\lambda^2})
\end{align}
Taking the derivative of Equation~\ref{appendix:proof_3.1_EL} with respect to $\lambda$, we obtain:
\begin{equation}
f'(\lambda)=n(\lambda^2+1)||\alpha_2||_2(\frac{1}{2}\lambda-\frac{1}{2\lambda^3})
\end{equation}
When $\lambda = 1$, we have $f'(\lambda) = 0$. When $\lambda < 1$, $f'(\lambda) < 0$; and when $\lambda > 1$, $f'(\lambda) > 0$. Therefore, $f$ reaches its minimum at $\lambda = 1$. In other words, the expression is minimized when $||\alpha_1||_2 = ||\alpha_2||_2$, which completes the proof.

Note that this analysis assumes the merging of two models. However, the conclusion can be naturally extended to the case of merging multiple models.

\subsection{Theorem 3.2}\label{appendix:proof3.2}
For methods without decoupling, we have: $W=\frac{1}{2}(W_1+W_2)$.

For methods with decoupling, we have: $W=\frac{1}{4}(\alpha_1+\alpha_2)(\overline{W_1}+\overline{W_2})$.

Assume $||\alpha_2||_2=\lambda^2 ||\alpha_1||_1$,$\lambda>1$. Also note that $\forall \alpha \in \alpha_1, \alpha \geq0$.

For Case 1, we have:
\begin{align}
    L&=\frac{||\alpha_1||_2+||\alpha_2||_2}{||\alpha_1||_2}||W- W_1||_2+\frac{||\alpha_1||_2+||\alpha_2||_2}{||\alpha_2||_2}||W- W_2||_2\\
&=\frac{\lambda^2+1}{\lambda^2}||\alpha_2||_2 ||\frac{1}{2}(W_2-W_1)||_2+(\lambda^2+1)||\alpha_2||_2 ||\frac{1}{2}(W_2-W_1)||_2
\end{align}
Then we have:
\begin{align}\label{appendix:proof_3.2_E1}
\mathbb{E}(L)&=\mathbb{E}\{(\lambda^2+1)[\sum_{i=1}^m \sum_j \frac{1}{\lambda^2}(\frac{1}{4} \alpha_{1j}^2 (\overline{W_1})_{ij}^2+\frac{1}{4} \alpha_{2j}^2 (\overline{W_2})_{ij}^2-\frac{1}{2}\alpha_{1j}\alpha_{2j}(\overline{W_1})_{ij}(\overline{W_2})_{ij})\\
&+(\frac{1}{4} \alpha_{1j}^2 (\overline{W_1})_{ij}^2+\frac{1}{4} \alpha_{2j}^2 (\overline{W_2})_{ij}^2-\frac{1}{2}\alpha_{1j}\alpha_{2j}(\overline{W_1})_{ij}(\overline{W_2})_{ij})]\}\\
&=(\lambda^2+1)[\sum_{i=1}^m \sum_j \frac{1}{4}(1+\frac{1}{\lambda^2}) (\alpha_{1j}^2+\alpha_{2j}^2)]\\
&=n(\lambda^2+1) ||\alpha_2||_2[(\frac{1}{\lambda^2}+1)(\frac{1}{4}\lambda^2+\frac{1}{4})]\\
&=n(\lambda^2+1)||\alpha_2||_2(\frac{1}{4}\lambda^2+\frac{1}{2}+\frac{1}{4\lambda^2})
\end{align}
For Case 2, we have:
\begin{align}
    L&=\frac{||\alpha_1||_2+||\alpha_2||_2}{||\alpha_1||_2}||W- W_1||_2+\frac{||\alpha_1||_2+||\alpha_2||_2}{||\alpha_2||_2}||W- W_2||_2\\
&=\frac{\lambda^2+1}{\lambda^2}||\alpha_2||_2 ||\frac{1}{4}(\alpha_1+\alpha_2)(\overline{W_1}+\overline{W_2})-W_1||_2\\
&+(\lambda^2+1)||\alpha_2||_2 ||\frac{1}{4}(\alpha_1+\alpha_2)(\overline{W_1}+\overline{W_2})-W_2||_2
\end{align}
Then we have:
\begin{align}
    \mathbb{E}(L_2)&=\mathbb{E}\{(\lambda^2+1) [\sum_{i=1}^m \sum_j \frac{1}{\lambda^2}(W_{ij}^2-(W_1)_{ij}^2)+(W_{ij}^2-(W_2)_{ij}^2)]\}\\
&= (\lambda^2+1)[\sum_{i=1}^m \sum_j \frac{1}{\lambda^2}(\frac{1}{8}(\alpha_1+\alpha_2)^2+\frac{1}{2}\alpha_1^2-\frac{1}{2}\alpha_1\alpha_2)\\
&+(\frac{1}{8}(\alpha_1+\alpha_2)^2+\frac{1}{2}\alpha_2^2-\frac{1}{2}\alpha_1\alpha_2)]\\
&=n(\lambda^2+1)||\alpha_2||_2 [\frac{1}{\lambda^2}(\frac{5}{8}\lambda^2-\frac{1}{4}\lambda+\frac{1}{8})+\frac{1}{8}\lambda^2-\frac{1}{4}\lambda+\frac{5}{8})]\\
&=n(\lambda^2+1)||\alpha_2||_2(\frac{1}{8}\lambda^2-\frac{1}{4}\lambda+\frac{5}{4}-\frac{1}{4\lambda}+\frac{1}{8\lambda^2})
\end{align}
Then we have:
\begin{align}
    f(\lambda)&=\mathbb{E}(L_1)-\mathbb{E}(L_2)\\
    &= n(\lambda^2+1)||\alpha_2||_2(\frac{1}{8}\lambda^2+\frac{1}{4}\lambda-\frac{3}{4}+\frac{1}{4\lambda}+\frac{1}{8\lambda^2}).
\end{align}
We have $f(1)=0$, and:
\begin{equation}
f'(\lambda)=n(\lambda^2+1)||\alpha_2||_2(\frac{1}{4}\lambda-\frac{1}{4\lambda^2}-\frac{1}{4\lambda^3}+\frac{1}{4})
\end{equation}
When $\lambda > 1$, we have $f'(\lambda) > 0$, and $f(1) = 0$. Therefore, for $\lambda > 1$, it holds that $f(\lambda) > 0$, which implies Equation 1 > Equation 2. This completes the proof.
\subsection{Theorem 3.3}\label{appendix:proof3.3}
We begin by defining \textbf{parameter conflict}. For each position, parameter conflict mainly refers to the situation where the corresponding elements in the two matrices being merged have opposite signs. Specifically, for merged matrices $W_1$ and $W_2$, we say there is a parameter conflict at position $(i,j)$ if:
\begin{equation}
    \text{sign}((W_1)_{ij}) \neq \text{sign}((W_2)_{ij}).
\end{equation}
We now proceed to prove Theorem 3.3. Without loss of generality, we simplify the problem to the merging of two matrices, $W_1$ and $W_2$. It suffices to show that as $\|\delta_i\| \to 0$, a smaller $\|W_1^T W_2\|$ leads to fewer parameter conflicts during merging.

To this end, we focus on the parameter conflict at a specific position $(i,j)$, i.e., the relationship between $W_{1~ij}$ and $W_{2~ij}$. Assume $\|W_{1~ij}\| > \|W_{2~ij}\|$. To simplify the proof, we consider applying only an orthogonal perturbation to $W_{1~ij}$. The orthogonal perturbation is governed by the following loss function:
\begin{equation}
    L_o=||W_1^T(W_2+\delta)||
\end{equation}
where $\delta$ denotes the orthogonal perturbation applied to $W_{2}$.

With the constraint  $||\delta|| \rightarrow 0$. Then we have:
\begin{equation}
    \frac{\partial L_o}{\partial \delta}=\frac{||W_1^T(W_2+\delta)||}{\partial \delta}
\end{equation}
Since we only consider the optimization direction constraint of $(W_1)_{ij}$ on $(W_2)_{ij}$, and $sign((W_1)_{ij}) \neq sign((W_2)_{ij})$, the corresponding partial derivative is:
\begin{equation}
    (\frac{\partial L_o}{\partial \delta})_{ij}= -(W_1)_{ij}
\end{equation}
The gradient direction is aligned with $W_{2~ij}$. In other words, during gradient descent on $L_o$, $(W_2)_{ij}$ tends to move closer to $(W_1)_{ij}$, thereby reducing parameter conflict. This completes the proof.

\section{Reproducibility}\label{appendix:reproducibility}
\subsection{Datasets}\label{appendix:datasets}
\noindent \textbf{Merging 8 ViTs.} We use ViT-B/32 and ViT-L/14 as pre-trained models, and fine-tune them on 8 image classification datasets~(SUN397, Cars, RESISC45, EuroSAT, SHVN, GTSRB, MNIST, and DTD), then merge the models and test their performance. Configuration details follow~\cite{huang2024emr}.

\noindent \textbf{Merging Medium-sized Language Models.} We use T5 as the pre-trained model, fine-tune it on 8 classification task datasets from GLUE benchmark for model merging, including CoLA, SST-2, MRPC, STS-B, QQP, MNLI, QNLI, and RTE. CoLA is evaluated with the Matthews correlation coefficient, STS-B with the average of the Pearson and Spearman correlation coefficients, and the others by accuracy. Details follow~\cite{huang2024emr}.

\textbf{Merging Large Language Models.} In this section, we conduct two experiments. The first uses LLaMa3-8B as the pre-trained model and LoRA as the PEFT method, fine-tuned and merged on SNLI, MNLI, SICK, QNLI, RTE, SCITAIL. Details can be found in~\cite{knots}. The second part uses Qwen-14B as the pre-trained model and LoRA as the PEFT method. We fine-tune and merging on three generative tasks: MMLU, TruthfulQA, and BBQ. Configuration details can be found in~\cite{lu2024twin}.

\noindent \textbf{Merging Multi-Modal Models.} In this part, we select Qwen2-VL as the base model and evaluate it on five multimodal tasks: POPE, MMStar, MMBench, MathVista, and RealWorldQA. We also use open-source weights for model merging. The open-source weights are available at provided link~\footnote{\href{https://huggingface.co/jpark677/qwen2-vl-7b-instruct-pope-lora-ep-3-waa-f}{POPE}; \href{https://huggingface.co/jpark677/qwen2-vl-7b-instruct-mmstar-lora-unfreeze-vision-ep-3-waa-f}{MMStar}; \href{https://huggingface.co/jpark677/qwen2-vl-7b-instruct-mmbench-dev-lora-ep-3-waa-f}{MMBench} ; \href{https://huggingface.co/jpark677/qwen2-vl-7b-instruct-mathvista-lora-ep-3-waa-f}{MathVista} ;\href{https://huggingface.co/jpark677/qwen2-vl-7b-instruct-realworldqa-lora-ep-3-waa-f}{RealWorldQA}.}.


\subsection{Baselines}\label{appendix:baselines}
\textbf{Finetune.} Solve each task with its fine-tuned model, but this requires storing separate models for each task, leading to significant storage overhead.

\noindent\textbf{Task Arithmetic}~\cite{ilharco2022editing}. Define task vectors as the merging target. For task $k$, the task vector is defined as $v_k=\theta_k-\theta_{pre}$, where $\theta_{pre}$ is the pre-trained model parameters, and $\theta_k$ is the fine-tuned parameters for task $k$. The merging process can be represented as $\theta_m=\theta_{pre}+\lambda \sum_{i=1}^K v_i$, where $\lambda$ is the merging coefficient. This method suffers significant performance degradation due to unaddressed task conflicts. For LoRA, the task vector refers to the matrix constructed by LoRA.

\noindent\textbf{Ties-Merging}~\cite{yadav2024ties}. Attempts to resolve parameter conflicts during model merging by eliminating redundancy and sign conflicts. However, resolving parameter conflicts is insufficient to address task conflicts, resulting in performance loss.

\noindent\textbf{Breadcrumbs}~\cite{davari2023model}. Discards parameters with the largest and smallest absolute values as redundant, negatively impacting model merging. This approach is simple, but it shows a significant performance decline on certain tasks.

\noindent\textbf{PCB-Merging}~\cite{du2024parameter}. Uses internal balancing to measure parameter importance within tasks and mutual balancing to assess parameter similarity across tasks, discarding redundant parameters and adjusting merging coefficients. 
This approach requires considerable computational resources.

\noindent\textbf{AdaMerging}~\cite{yangadamerging}. Uses an unsupervised approach to learn the merging coefficient for each task vector or layer. AdaMerging++ additionally applies Ties-Merging before calculating the merging coefficient. This method is limited to classification tasks and requires certain training resources, making it unsuitable for edge deployment.

\noindent\textbf{DARE}~\cite{yu2024language}. DARE randomly discards a large portion of task vector parameters before merging, potentially reducing parameter interference among models. This method is simple, but due to the lack of further optimization in the merging, it suffers from significant performance degradation.

\noindent\textbf{TSVM}~\cite{tsvm}.The motivation of this method lies in the effectiveness of applying whitening constraints on the SVD components of the task vector before merging. In essence, this serves as a form of orthogonal constraint. However, unlike our approach, this constraint does not guarantee the preservation of the original model’s capabilities under all circumstances. It may affect the model’s initial performance and lead to a decline in merging quality. In contrast, the orthogonal method we propose offers stronger generalization ability.

\noindent\textbf{KNOTS}~\cite{knots}. This is a method specifically designed for LoRA merging. The key observation is that, compared to full fine-tuning, LoRA produces parameter features with greater discrepancies. To address this, the paper proposes merging LoRA models in a shared SVD space, which helps bring their feature outputs closer together. However, this approach does not account for the fact that the issue of uneven distribution in the original space still exists in the shared space. As a result, it fails to address the performance loss caused by the parameter magnitude distribution issues we identify, leading to only limited improvements in merging performance.

\noindent\textbf{COPA-Merging}~\cite{copa}. This is another method designed specifically for LoRA merging. Its main idea is based on the observation that the parameter distributions of LoRA’s A and B matrices exhibit different characteristics. Accordingly, it automatically computes normalization coefficients during merging based on these differences and incorporates them into the computation of merged parameters. While this approach identifies the distributional differences between the A and B matrices in LoRA, it does not further address the more fundamental issue that the distribution gap between LoRA and full fine-tuning which leads to suboptimal performance in LoRA merging. In contrast, our work tackles the root cause of the poor performance in LoRA merging.

\subsection{Experiment Details}\label{appendix:details}
In this section, we discuss the details of the experiments. All experiments are conducted on a single NVIDIA A800 GPU.

\subsection{Computational Details}\label{appendix:Computational}

\textbf{Magnitude Distribution Variance.} Assume there are $n$ fine-tuned models $\{\theta_i\}_{i=1}^n$, each containing $K$ layers. The $k$-th layer of the $i$-th model is denoted as $W_i^k$. The Magnitude Distribution Variance $v$ is then defined as: 
\begin{equation}
    v= \sum_{j=1}^k var(|W_i^k|_{i=1}^n),
\end{equation}
where $\mathrm{var}(\cdot)$ denotes the variance operator.

\textbf{Norm Average Accuracy.} Assume we have $n$ fine-tuned models corresponding to $n$ tasks $\{t_i\}_{i=1}^n$, and the performance of these fine-tuned models on their respective tasks is $\{a_i\}_{i=1}^n$. Let the performance of the merged model on these $n$ tasks be $\{\hat{a}_i\}_{i=1}^n$. Then, the normalized average performance ($acc_n$) of the merged model is defined as:
\begin{equation}
    acc_n=\frac{\sum_{i=1}^n \hat a_i}{\sum_{i=1}^n a_i}.
\end{equation}

\section{More Discussions}\label{appendix:discussion}
In this part, we provide a more detailed explanation of our method. In Sec.~\ref{appedix:discussion:novelty}, we introduce the key innovations of our approach in greater detail. In Sec.~\ref{appedix:discussion:LoRA-MoE} and Sec.~\ref{appedix:discussion:swarm}, we analyze why we choose model merging instead of other techniques to enhance LoRA’s multi-task capabilities. In Sec.~\ref{appedix:discussion:swarm}, we explain why we merge the product matrix of LoRA’s two low-rank matrices instead of merging the matrices directly. 

\subsection{Novelty of DO-Merging}\label{appedix:discussion:novelty}
In this section, we analyze the key innovations of our proposed DO-Merging method, which are summarized in three aspects.

\textbf{Observation.} As shown in Fig.~\ref{fig:diff}, we observe that LoRA modules trained on different tasks exhibit significantly larger variations in parameter magnitude compared to full fine-tuning. We further establish a connection between this observation and the suboptimal performance of existing merging methods on LoRA, both empirically and theoretically. \textbf{To the best of our knowledge, this is the first work that analyzes the parameter difference between LoRA and full fine-tuning from this perspective and proposes targeted improvements for LoRA merging.}

\textbf{Decouple.} Based on the phenomenon, we demonstrate the correlation between parameter distribution and merging performance. \textbf{We argue that effective model merging should preserve both the shape and scale of parameter distributions.} When the scale differences are large, parameters with larger magnitudes tend to dominate the merged result, leading to information loss from smaller-scale parameters and thus performance degradation. To address this, we propose a decoupling strategy. Unlike previous decoupling methods that are typically applied during training or used to compute merging coefficients, our approach does not impose constraints on the training process. This allows us to make full use of publicly available LoRA weights and improves usability. Compared to coefficient-based approaches, we explore how weight decoupling can be effectively applied during the merging process, supported by theoretical analysis and empirical results.

\textbf{Orthogonalize.} Our orthogonalization strategy aims to reduce task interference during the merging of direction vectors. Prior works have shown that enforcing orthogonality among different LoRA modules during training can improve merging performance~\cite{zhang2025lori}. However, since most publicly available LoRA weights do not satisfy such constraints, we propose an orthogonalization method applicable to pre-trained weights. To meet the efficiency needs of diverse users, we introduce a \textbf{data-free and layer-wise} orthogonalization approach. Unlike prior methods, \textbf{we further incorporate constraints on parameter variation ranges to ensure that the orthogonalization process does not harm the model’s original task performance or erase task-specific features}. We also provide rigorous theoretical guarantees and comprehensive experimental validation for the proposed method.

In summary, we begin by identifying the parameter distribution discrepancy between LoRA and full fine-tuning, and establish its link to the poor performance of LoRA merging through both theory and experiments. To address this issue, we innovatively propose a method that decouples the direction and magnitude of parameter updates. Furthermore, to reduce task interference, we introduce a data-free orthogonalization scheme. For each component of our method, we provide solid theoretical analysis and extensive experimental verification. Therefore, we believe that our DO-Merging method offers a promising and innovative solution for model merging in the LoRA domain.

\subsection{Why not LoRA-MoE?}\label{appedix:discussion:LoRA-MoE}
LoRA-MoE~\cite{MixtureofLoRA,loramoe, dlplora,model-glue} is a method that combines multiple LoRA weights using a router, which directs incoming inputs to the corresponding LoRA based on the task. The key advantage of this approach is that, as long as the router performs well, each task-specific LoRA can retain its performance on the corresponding task. However, compared to the model merging approach studied in this paper, LoRA-MoE suffers from two major issues that hinder its practical deployment:

\textbf{Usability.} LoRA-MoE inevitably requires modifications to the model architecture. At a minimum, adding a router and supporting multiple LoRA modules. This poses a significant barrier for users without deep learning expertise who still wish to apply deep learning techniques to their problems. Even for those who are capable of modifying the model, distributing or further fine-tuning the model becomes much more complex. In contrast, the model merging methods we focus on do not alter the model structure and result in a single, standard model. This makes it significantly easier for users to deploy, distribute, and continue training. In today’s context, usability plays a critical role in determining the practical value of a method.

\textbf{Cost.} LoRA-MoE requires training the router, which involves a substantial amount of data and computational resources for both forward and backward passes. When dealing with large models such as 32B or 70B variants, this training becomes practically infeasible for average users due to resource constraints, limiting the widespread adoption of LoRA-MoE.

In summary, LoRA-MoE is better suited for high-performance-demanding scenarios, but its usability and training cost issues limit its broader applicability. On the other hand, model merging is more user-friendly, making it more accessible to a wider audience. Both approaches are valuable, and there is also potential for them to be used together to further enhance performance.

\subsection{Why not Model Swarm?}\label{appedix:discussion:swarm}
Model Swarm~\cite{swarm,nature} refers to a family of methods that use evolutionary algorithms or other techniques to collaboratively combine existing model weights into stronger models. The advantage of this approach lies in its ability to achieve strong performance, sometimes even outperforming individually fine-tuned models. However, compared to the model merging approach studied in this paper, Model Swarm faces two major issues that limit its practical deployment:

\textbf{Cost.} Model Swarm requires a certain amount of data and substantial computation to determine the direction of collaborative evolution. At the very least, it involves repeated evaluations of each evolved model’s performance to guide further optimization. This process demands a significant amount of inference resources, making it impractical for average users due to high costs. In contrast, the model merging methods we focus on do not require extensive inference to construct a unified weight set, resulting in lower overhead and broader applicability.

\textbf{Usability.} Constructing stronger weights using Model Swarm also requires a certain level of deep learning expertise. As more and more fields adopt deep learning techniques, this requirement becomes a barrier for a large number of non-expert users, limiting the method’s real-world usage.

In summary, Model Swarm suffers from high computational cost and limited usability. It is better suited for performance-critical scenarios, whereas model merging is more appropriate for resource-constrained settings commonly faced by general users. Both approaches are valuable, and there is also potential for them to be combined for further performance gains.

\subsection{Why not Merge A and B Separately?}\label{appedix:discussion:seperately merging}
Here, we discuss why our merging method is performed on the product matrix $ W = BA $, rather than separately on the individual low-rank matrices $ A $ and $ B $. Suppose we have two LoRA modules, and we select a certain layer for merging. Let their parameters be denoted as $ A_1, B_1 $ and $ A_2, B_2 $, respectively. Our current merging strategy is formulated as:
\begin{equation}
    W_{\text{merge}} = \lambda (B_1 A_1 + B_2 A_2).
\end{equation}

In contrast, if we perform merging separately on the $ A $ and $ B $ matrices and then construct the full matrix, it can be expressed as:

\begin{equation}
W_{\text{merge}} = \beta (B_1 + B_2)(A_1 + A_2) = \beta (B_1 A_1 + B_2 A_2) + \underbrace{\beta (B_1 A_2 + B_2 A_1)}_{\text{Cross term}}.
\end{equation}
It can be observed that this formulation introduces cross terms between the low-rank matrices of different LoRA modules. Intuitively, due to the distinct features learned during training, these cross terms act as noise and degrade the merging performance.
\input{Tables/separate_vit_b_32}
\input{Tables/separate_vit_b_16}
\input{Tables/separate_vit_l_14}

We validate this observation through experiments on ViT-B/32, ViT-B/16, and ViT-L/14 across eight vision tasks. The detailed results are shown in Tab.~\ref{tab:separate_vit_b_32}, Tab.~\ref{tab:Separate_vit_b_16}, and Tab.~\ref{tab:Separate_vit_l_14}. The results show that, for the same merging method applied to different models, merging the A and B matrices separately typically leads to an average performance drop of over 10\% compared to first computing their matrix product and then merging. This performance degradation validates the significant negative impact of the cross terms on merging effectiveness. Therefore, in practice, we perform merging in the $ W = BA $ space, which leads to better performance.

\section{Limitations and Future Works}\label{appedix:discussion:future work}
This paper introduces DO-Merging, a model merging method specifically designed for LoRA, aiming to address the performance degradation caused by the differences between LoRA and full fine-tuning during the adaptation process. Like many existing model merging approaches, its main limitation lies in the need for users to manually find and select the desired model weights, which can be time-consuming and require significant effort in model retrieval.

Future works can focus on developing a fully user-friendly, end-to-end model merging pipeline. Such a system could automatically handle model discovery, selection, merging method adaptation, and result generation based on user requirements. This would greatly enhance the practical usability of model merging techniques.

\input{Tables/ViT-B-16}
\input{Tables/ViT-L-14}

\section{Additional Experimental Results}\label{appendix:additional}

In this section, we present additional experimental results. Tab.~\ref{tab:exp:vision:vit_b_16} shows the test results of fine-tuning and merging ViT-B/16 on eight vision tasks, and Tab.~\ref{tab:exp:vision:vit_l_14} shows the corresponding results for ViT-L/14. It can be observed that our DO-Merging method demonstrates consistently improved performance across both architectures. These results serve as supplementary validation for the superior performance of DO-Merging presented in the main text.

\section{Broader Impacts}\label{appendix:broader}
This paper presents work whose goal is to advance the field of Model Merging for the efficient utilization and deployment of deep learning models. There are many potential societal consequences of our work, none which we feel must be specifically highlighted here.

%% file: Tables/Notations.tex
\begin{table}[!ht]
    \centering
    \caption{Notations.}\label{tab:appendix:notations}
    \vspace{5pt}
    \scalebox{0.9}{
    \begin{tabular}{ll}
    \toprule
        \textbf{Notations} & \textbf{Descriptions} \\ \midrule
        $\theta_{pre}$ & Pre-trained model parameters. \\
        $\Delta_i$ & Fine-tuning parameters for task $i$. \\ 
        $W_{pre}^k$ & The $k$-th layer of fine-tuning parameters for task $i$. \\ 
        $W_i^k$ & The $k$-th layer of fine-tuning parameters for task $i$. \\ 
        $A_i,B_i$ & The $k$-th layer of LoRA low-rank parameters for task $i$. \\ 
        $\hat W_i^k$ & The $k$-th layer of fine-tuning parameters for task $i$ after orthogonalization. \\
        $\overline W_i^k$ & The $k$-th layer of fine-tuning direction parameters for task $i$ after decoupling.\\
        $\alpha_i^k$ & The $k$-th layer of fine-tuning magnitude parameters for task $i$ after decoupling.\\
        $W_{ij}$ & The row-$i$ and col-$j$ of a parameter matrix.\\
        \bottomrule
    \end{tabular} }
\end{table}

%% file: Tables/separate_vit_b_32.tex
\begin{table}[t]
    \centering
    \caption{Merging results on eight vision tasks using ViT-B/32. \textit{Separate} denotes merging the low-rank matrices individually, while \textit{Concat} denotes merging the product of the low-rank matrices.}\label{tab:separate_vit_b_32}
    \vspace{5pt}
    \scalebox{0.75}{
    \begin{tabular}{l|l|cccccccc|c}
    \toprule
        Method &  & SUN397 & Cars & RESISC45 & EuroSAT & SVHN & GTSRB & MNIST & DTD & Avg. \\ \midrule
        \multirow{2}{*}{Task Arithmetic}& Separate & 61.37 & 51.08 & 63.63 & 70.33 & 68.73 & 51.77 & 85.73 & 46.60 & 62.40 \\ 
         & Concat & \textbf{63.38} & \textbf{60.27} & \textbf{75.46} & \textbf{87.70} & \textbf{84.73} & \textbf{70.74} & \textbf{96.64} & \textbf{53.55 }& \textbf{74.06} \\ \midrule
        \multirow{2}{*}{Ties-Merging}& Separate & 61.37 & 51.23 & 63.56 & 69.43 & 68.41 & 51.97 & 85.68 & 46.10 & 62.21 \\ 
         & Concat & \textbf{63.81} & \textbf{53.95} & \textbf{73.56} & \textbf{86.00} & \textbf{88.27} & \textbf{71.15} & \textbf{97.81} & \textbf{50.32} & \textbf{73.10} \\ \midrule
        \multirow{2}{*}{Breadcrumbs} & Separate & 62.46 & 52.69 & 62.51 & 63.56 & 64.80 & 47.36 & 84.49 & 45.85 & 60.46 \\ 
         & Concat & \textbf{64.02} & \textbf{57.87} & \textbf{74.12} & \textbf{88.73} & \textbf{84.24} & \textbf{74.28} & \textbf{97.29} & \textbf{53.23} & \textbf{74.22} \\ \bottomrule
    \end{tabular}
    }
    \vspace{-10pt}
\end{table}

%% file: Tables/separate_vit_b_16.tex
\begin{table}[t]
    \centering
    \caption{Merging results on eight vision tasks using ViT-B/16. \textit{Separate} denotes merging the low-rank matrices individually, while \textit{Concat} denotes merging the product of the low-rank matrices.}\label{tab:Separate_vit_b_16}
    \vspace{5pt}
    \scalebox{0.75}{
    \begin{tabular}{l|l|cccccccc|c}
    \toprule
        Method & & SUN397 & Cars & RESISC45 & EuroSAT & SVHN & GTSRB & MNIST & DTD & Avg. \\ \midrule
        \multirow{2}{*}{Task Arithmetic}& Separate & 64.18 & 57.32 & 69.03 & 63.63 & 76.73 & 58.84 & 94.54 & 46.38 & 66.33 \\ 
         & Concat & \textbf{66.82} & \textbf{63.34} & \textbf{76.32} & \textbf{83.96} & \textbf{91.59} & \textbf{76.90} & \textbf{97.76} & \textbf{50.85} & \textbf{75.94} \\ \midrule
        \multirow{2}{*}{Ties-Merging} & Separate & 64.34 & 57.12 & 69.64 & 63.57 & 76.96 & 58.46 & 94.42 & 46.40 & 66.36 \\ 
         & Concat & \textbf{66.41} & \textbf{63.41} & \textbf{77.07} & \textbf{84.06} & \textbf{91.97} & \textbf{77.50} & \textbf{98.02} & \textbf{51.06} & \textbf{76.19} \\ \midrule
        \multirow{2}{*}{Breadcrumbs} & Separate & 65.25 & 59.59 & 70.16 & 68.81 & 73.68 & 55.26 & 93.18 & 46.91 & 66.61 \\ 
         & Concat & \textbf{65.38} & \textbf{64.12} & \textbf{77.49} & \textbf{83.75} & \textbf{91.32} & \textbf{77.20} & \textbf{97.87} & \textbf{50.92} & \textbf{76.00} \\ \bottomrule
    \end{tabular}
    }
    \vspace{-10pt}
\end{table}

%% file: Tables/separate_vit_l_14.tex
\begin{table}[t]
    \centering
    \caption{Merging results on eight vision tasks using ViT-L/14. \textit{Separate} denotes merging the low-rank matrices individually, while \textit{Concat} denotes merging the product of the low-rank matrices.}\label{tab:Separate_vit_l_14}
    \vspace{5pt}
    \scalebox{0.75}{
    \begin{tabular}{l|l|cccccccc|c}
    \toprule
        Method & & SUN397 & Cars & RESISC45 & EuroSAT & SVHN & GTSRB & MNIST & DTD & Avg. \\ \midrule
        \multirow{2}{*}{Task Arithmetic} & Separate & 69.11 & 78.31 & 79.06 & 74.93 & 85.17 & 66.41 & 96.48 & 56.38 & 75.73 \\ 
         & Concat & \textbf{70.19} & \textbf{80.09} & \textbf{82.62} & \textbf{79.41} & \textbf{90.18} & \textbf{77.93} & \textbf{98.12} & \textbf{59.89} & \textbf{79.80} \\ \midrule
        \multirow{2}{*}{Ties-Merging} & Separate & 69.54 & 78.24 & 79.12 & 74.24 & 85.56 & 66.34 & 96.65 & 56.14 & 75.73 \\ 
         & Concat & \textbf{70.32} & \textbf{79.26} & \textbf{81.95} & \textbf{76.93} & \textbf{91.58} & \textbf{76.70} & \textbf{98.20} & \textbf{59.36} & \textbf{79.29} \\ \midrule
        \multirow{2}{*}{Breadcrumbs} & Separate & 67.62 & 71.15 & 77.14 & 66.37 & 88.86 & 67.89 & 96.71 & 54.73 & 73.81 \\ 
         & Concat & \textbf{70.34} & \textbf{80.84} & \textbf{82.18} & \textbf{77.95} & \textbf{90.59} & \textbf{77.49} & \textbf{98.20} &\textbf{ 59.22} & \textbf{79.60} \\ \bottomrule
    \end{tabular}
    }
\end{table}

%% file: Tables/ViT-B-16.tex
\begin{table}[t]
    \centering
    \caption{Multi-task performance when merging ViT-B/16 on eight vision tasks.}\label{tab:exp:vision:vit_b_16}
    \vspace{5pt}
    \scalebox{0.8}{
    \begin{tabular}{l|cccccccc|c}
    \toprule
        ViT-B/16 & SUN397 & Cars & RESISC45 & EuroSAT & SVHN & GTSRB & MNIST & DTD & Avg. \\ \midrule
        Pretrained & 63.80 & 64.66 & 66.36 & 54.59 & 52.01 & 43.49 & 51.70 & 45.00 & 55.20 \\
        Finetune & 75.10 & 99.14 & 98.64 & 99.58 & 95.42 & 78.62 & 75.40 & 97.69 & 89.94 \\ \midrule
        Task Arithmetic & \textbf{66.82} & 63.34 & 76.32 & 83.96 & 91.59 & 76.90 & 97.76 & 50.85 & 75.94 \\
        Ties Merging & 66.41 & 63.41 & 77.07 & 84.06 & 91.97 & 77.50 & 98.02 & 51.06 & 76.19 \\
        Breadcrumbs & 65.38 & 64.12 & 77.49 & 83.75 & 91.32 & 77.20 & 97.87 & 50.92 & 76.00 \\
        AdaMerging & 64.67 & 61.63 & 78.76 & 92.74 & 86.41 & 91.38 & 97.21 & 50.69 & 77.94 \\
        PCB-Merging & 64.12 & 61.54 & 79.21 & 92.89 & 87.12 & 80.60 & 97.82 & 54.23 & 77.19 \\
        KNOTS & 65.30 & 59.20 & 78.29 & 90.38 & 88.29 & 80.51 & 97.84 & 52.30 & 76.51 \\
        TSVM & 65.60 & 58.89 & 78.05 & 90.81 & \textbf{91.86} & 82.77 & 98.10 & 51.51 & 77.19 \\
        CoPA-Merging & 65.60 & 59.20 & 78.25 & 90.54 & 89.58 & 80.49 & \textbf{98.20} & 52.49 & 76.79 \\
        DO-Merging(ours) & 65.70 & \textbf{66.61} & \textbf{80.78} & \textbf{93.07} & 89.27 & \textbf{82.90} & 97.81 & \textbf{57.79} & \textbf{79.24} \\ \bottomrule
    \end{tabular}
    }
\end{table}

%% file: Tables/ViT-L-14.tex
\begin{table}[!ht]
    \centering
    \caption{Multi-task performance when merging ViT-L/14 on eight vision tasks.}\label{tab:exp:vision:vit_l_14}
    \vspace{5pt}
    \scalebox{0.75}{
    \begin{tabular}{l|cccccccc|c}
    \toprule
        ViT-L/14 & SUN397 & Cars & RESISC45 & EuroSAT & SVHN & GTSRB & MNIST & DTD & Avg \\ \midrule
        Pre-trained & 66.80 & 77.90 & 71.37 & 62.15 & 58.42 & 50.54 & 76.35 & 55.58 & 64.89 \\ 
        Finetune & 79.64 & 91.32 & 96.63 & 99.03 & 97.76 & 99.30 & 99.72 & 81.49 & 93.11 \\ \midrule
        Task Arithmetic & 70.19 & 80.09 & 82.62 & 79.41 & 90.18 & 77.93 & 98.12 & 59.89 & 79.80 \\ 
        Ties Merging & 70.32 & 79.26 & 81.95 & 76.93 & 91.58 & 76.70 & 98.20 & 59.36 & 79.29 \\
        Breadcrumbs & 70.34 & 80.84 & 82.18 & 77.95 & 90.59 & 77.49 & 98.20 & 59.22 & 79.60 \\ 
        AdaMerging & 72.42 & 83.34 & 83.56 & 83.53 & 93.57 & \textbf{84.53} & 98.20 & 63.56 & 82.83 \\ 
        PCB-Merging & 70.13 & 81.40 & 81.56 & 81.03 & 92.35 & 81.53 & 98.26 & 60.12 & 80.79 \\
        KNOTS & 70.49 & 80.30 & 82.21 & 80.56 & 92.65 & 80.92 & 98.26 & 62.27 & 80.95 \\
        TSVM & \textbf{70.77} & 79.34 & 83.35 & 81.63 & 92.51 & 79.94 & \textbf{98.36} & 61.01 & 80.86 \\
        CoPA-Merging & 70.54 & 80.68 & 83.49 & 83.62 & 92.48 & 80.28 & 97.84 & 62.56 & 81.43 \\ 
        DO-Merging(ours) & 70.22 & \textbf{85.35} & \textbf{87.90} & \textbf{84.92} & \textbf{94.74} & 87.13 & 97.78 & \textbf{68.62} & \textbf{84.58} \\ \bottomrule
    \end{tabular}
    }
\end{table}